\def\year{2022}\relax
\documentclass[letterpaper]{article} % DO NOT CHANGE THIS
\usepackage{aaai22}  % DO NOT CHANGE THIS
\usepackage{times}  % DO NOT CHANGE THIS
\usepackage{helvet} % DO NOT CHANGE THIS
\usepackage{courier}  % DO NOT CHANGE THIS
\usepackage[hyphens]{url}  % DO NOT CHANGE THIS
\usepackage{graphicx} % DO NOT CHANGE THIS
\usepackage[utf8]{inputenc} % allow utf-8 input
\usepackage[T1]{fontenc}    % use 8-bit T1 fonts

\usepackage{booktabs}       % professional-quality tables
\usepackage{amsfonts}       % blackboard math symbols
\usepackage{nicefrac}       % compact symbols for 1/2, etc.
\usepackage{microtype}      % microtypography
\usepackage{xcolor}         % colors
\usepackage{subcaption}

\usepackage{soul}
\usepackage{url}
\usepackage[hidelinks]{hyperref}
\usepackage{amsmath}
\DeclareCaptionStyle{ruled}{labelfont=normalfont,labelsep=colon,strut=off} % DO NOT CHANGE THIS
\usepackage[ruled,vlined,linesnumbered]{algorithm2e}
\usepackage{textcomp}
\usepackage{xcolor}
\usepackage{array}
\usepackage{setspace}
\usepackage{caption}
\usepackage[multiple]{footmisc}
\usepackage{multirow}
\usepackage{natbib}  % DO NOT CHANGE THIS AND DO NOT ADD ANY OPTIONS TO IT
\usepackage{caption} % DO NOT CHANGE THIS AND DO NOT ADD ANY OPTIONS TO IT
\title{GearNet: Stepwise Dual Learning for Weakly Supervised Domain Adaptation}
\author {
    Renchunzi Xie\textsuperscript{\rm 1}, Hongxin Wei\textsuperscript{\rm 1}\thanks{Corresponding Author}, Lei Feng\textsuperscript{\rm 2} and Bo An\textsuperscript{\rm 1} 
}

\affiliations{
    \textsuperscript{\rm 1} School of Computer Science and Engineering, Nanyang Technological University, Singapore \\\textsuperscript{\rm 2}College of Computer Science, Chongqing University, China \\
    XIER0002@e.ntu.edu.sg, hongxin001@e.ntu.edu.sg, lfeng@cqu.edu.cn, boan@ntu.edu.sg
}
\begin{document}

\maketitle

\begin{abstract}
This paper studies a \emph{weakly supervised domain adaptation} (WSDA) problem, where we only have access to the source domain with \emph{noisy labels}, from which we need to transfer useful information to the unlabeled target domain. Although there have been a few studies on this problem, most of them only exploit \emph{unidirectional} relationships from the source domain to the target domain. In this paper, we propose a universal paradigm called GearNet to exploit \emph{bilateral} relationships between the two domains. Specifically, we take the two domains as different inputs to train two models alternately, and a symmetrical Kullback-Leibler loss is used for selectively matching the predictions of the two models in the same domain. %By equipping with our GearNet strategy, all existing WSDA models in the training process can learn hidden domain relationships from the source-to-target and target-to-source patterns, and reach a continuously incremental performance by mutual noise reduction between the two domains. 
% By equipping with our GearNet, existing WSDA models are able to exploit more hidden domain relationships and alleviate the label noise by following the source-to-target and target-to-source training patterns.
This interactive learning schema enables implicit label noise canceling and exploit correlations between the source and target domains. Therefore, our GearNet has the great potential to boost the performance of a wide range of existing WSDA methods.
Comprehensive experimental results show that the performance of existing methods can be significantly improved by equipping with our GearNet. 
% potentially positive information
% significantly outperforms state-of-the-art counterparts.

%  we can train the current model to reach a better performance based on the achievement of the last model. In the end, an improvement chain can be constructed to enhance the predictive accuracy of the target domain step by step.
% This  makes the training process  the performance of target inputs maintain a high level, so that the target inputs of a model are more likely to provide useful information with the source inputs of the same model in turn. In the end, an improvement chain can be constructed to enhance the predictive accuracy of the target domain step by step. Extensive experimental results show that GearNet significantly outperforms state-of-the-art counterparts.
% the the target outputs of the current training model with the source outputs of the last model. Meanwhile, a symmetrical Kullback-Leibler (KL) loss is calculated to match the target outputs of the current training model with the source outputs of the last model, 
% so that the target accuracy can be maintained at a high level, and in turn the target inputs are more likely to provides useful information with the source inputs in the same model. Extensive experimental results show that GearNet can improve the target accuracy step by step, and significantly outperforms state-of-the-art counterparts.
\end{abstract}
% We treat the two domains as different inputs and train two models alternately,

\section{Introduction}
% \emph{Weakly supervised domain adaptation} (WSDA) aims to learn a discriminative predictor for the unlabeled target domain when the training data from the source domain present label noise as well as distribution shifts \cite{shu2019transferable}. Its setting is more realistic compared with Unsupervised domain adaptation (UDA) which only suffers from domain shifts \cite{combes2020domain, zhang2013domain, pan2009survey}, since collecting precisely labeled data for the source domain is usually expensive and time-consuming \cite{frenay2013classification, ghosh2017robust}.

In the problem of domain adaptation, we aim to train classifiers for data from the target domain by leveraging auxiliary data sampled from related but different source domains \cite{combes2020domain, zhang2013domain, pan2009survey, 9616392, What_Transferred_Dong_CVPR2020}. Most of the existing domain adaptation studies assume that the source domains are clean datasets with accurate annotations. However, it is usually expensive and time-consuming to collect such large-scale and correctly labeled datasets in some real-world scenarios \cite{frenay2013classification, ghosh2017robust}. To alleviate this problem, an increasing number of researchers started to investigate \emph{weakly supervised domain adaptation} (WSDA), where only source domain data with noisy labels and unlabeled target domain data are available.

To improve the model robustness against label noise from the source domain, some WSDA algorithms \cite{shu2019transferable,liu2019butterfly,yu2020label} were developed to specially reduce the negative impact of label noise while minimizing the distribution discrepancy of two domains. For example, TCL \cite{shu2019transferable} selects clean and transferable samples from the source domain guided by a transferable curriculum and Butterfly \cite{liu2019butterfly} picks small-loss samples from both the source domain and target domain. Similarly, DCIC \cite{yu2020label} emphasizes clean and transferable source samples by an estimated transition matrix. Although these methods achieve acceptable performance, they only exploit the supervision information from the source domain to prevent the model from overfitting to label noise, and then transfer the learned information from the source domain to the target domain. In other words, these methods only exploit \emph{unidirectional} relationships from the source domain to the target domain. However, if we further consider exploiting the pseudo supervision information from the unlabeled target domain, the relationships between the two domains are exploited in a \emph{bilateral} way. Consequently, richer supervision information could be discovered for combating the label noise and the gap between the two domains would be narrowed. To the best of our knowledge, we are the first to explore the benefit of utilizing pseudo supervision knowledge from the target domain in improving the robustness against noisy labels from the source domain.

 This paper proposes the first universal paradigm to exploit bilateral relationships between the source domain and the target domain.
%  Because it is non-trivial to obtain accurate supervised information from the target domain due to the absence of the target labels under the setting of domain adaptation, we utilize pseudo labels generated by the source-domain-training model instead. 
Specifically, we train two models on the two domains respectively in an alternate manner, and the whole training process consists of four main steps: 1) training model A on source domain data with noisy labels, 2) using model A to generate pseudo labels for target domain data, 3) training model B on pseudo-labeled target domain data with regularization on the consistency of source-domain class posteriors of model B and model A, 4) training model A on labeled source domain data with regularization on the consistency of target-domain class posteriors of model A and model B. We iterate from step 2 to step 4 multiple times until the training process stops. In this way, the pseudo supervision information in the target domain can be discovered and leveraged to improve the model robustness against label noise from source domain. It is worth noting that our proposed paradigm is a general WSDA framework to enhance the model robustness. Therefore, it can be easily incorporated into existing WSDA algorithms for further improving the performance of those methods.

 To verify the effectiveness of our proposed GearNet, we conduct extensive experiments on widely used benchmark datasets, and experimental results demonstrate that our GearNet can significantly improve the performance of existing robust methods. 

\section{Related Work}
\noindent\textbf{Unsupervised Domain Adaptation.}\quad \emph{Unsupervised domain adaptation} (UDA) has gained considerable interests in many practical applications recently \cite{shao2014transfer, hoffman2018cycada, hoffman2014lsda, ghafoorian2017transfer, kamnitsas2017unsupervised, wang2015transfer, blitzer2006domain, DBLP:journals/tnn/FangLLXZ21, DBLP:conf/NeuriPS/FangLLL021}, which aims to learn a model on data from the labeled source domain and transfer the learned information to a new unlabeled domain with distribution shift \cite{pan2009survey}. The key to the success of UDA is to learn a latent domain-invariant representation by minimizing the difference between the two domains (i.e., domain discrepancy) with certain criteria, such as maximum mean discrepancy \cite{pan2010domain}, Kullback-Leibler divergence \cite{zhuang2015supervised}, central moment discrepancy \cite{zellinger2017central}, and Wasserstein distance \cite{lee2017minimax}. Besides, some studies utilized the domain discriminator in an adversarial manner to minimize the domain discrepancy, like domain-adversarial neural network \cite{ganin2016domain} and Adversarial discriminative domain adaptation \cite{tzeng2017adversarial}. More recently, self-training based methods \cite{chen2020self, zou2019confidence} have been proposed for UDA, which are based on the motivation that the domain adaptation process uses the target label information estimated by the source-domain-training model to enhance itself. However, those methods require the assumption that all labels in the source domain are correct, which is difficult to satisfy in the real world. Therefore, it is of great significance for us to develop specially designed learning methods for UDA with label noise in the source domain (i.e., weakly supervised domain adaptation).
\vspace{0.1cm}

\noindent\textbf{Weakly Supervised Domain Adaptation.}\quad WSDA considers both the UDA problem and the label noise issue, which is more common in practical scenarios. There have been several studies \cite{shu2019transferable,tzeng2017adversarial,liu2019butterfly} to address the WSDA problem by training domain adaptation models with sample reweighting. For example, TCL \cite{shu2019transferable} selects clean and transferable source samples to train a neural network that has the same structure as DANN \cite{tzeng2017adversarial}; Butterfly \cite{liu2019butterfly} picks clean samples from both the source domain and the target domain while sharing the shallow layers of two Co-teaching models \cite{han2018co} for domain adaptation. DCIC \cite{yu2020label} emphasizes clean and transferable source data to construct a denoising maximum mean discrepancy \cite{pan2010domain} loss. Despite the effectiveness of these methods, they only exploit the supervision information from the source domain and regrettably ignore the potential supervision information in the target domain.
\vspace{0.1cm}
%to prevent the model from overfitting to label noise, and then transfer the learned information from the source domain to the target domain. In other words, these methods only exploit unidirectional relationships from the source domain to the target domain.
%But those works only use supervised information from the source domain to improve the model robustness, where the predictor trained by the de-noisy source data is still unreliable for the other domain.

\noindent\textbf{Learning with Noisy Labels.}\quad A wide range of algorithms have been proposed to improve the model robustness against label noise in the training data \cite{zhang2016understanding, han2019deep, menon2015learning, DBLP:conf/icml/FangLLL021}. Early studies \cite{goldberger2016training, patrini2017making} learn a robust model by estimating the label transition matrix to fit the noisy labels. However, They can only achieve mediocre performance, as it is non-trivial to obtain a high-quality estimation of noise rates. Recently, training with sample reweighting has became a popular research direction to handle label noise \cite{jiang2018mentornet, ren2018learning}, where reliable and noiseless data are emphasized during the training process. Another promising direction is to design noise-robust loss functions, such as mean absolute error \cite{ghosh2017robust}, generalized cross entropy loss \cite{zhang2018generalized}, and Taylor cross entropy loss \cite{feng2020can}. The above methods for learning with noisy labels have provided many inspirations for WSDA methods to combat label noise.
\begin{figure}[t]
    \centering
     \scalebox{.65}{\includegraphics{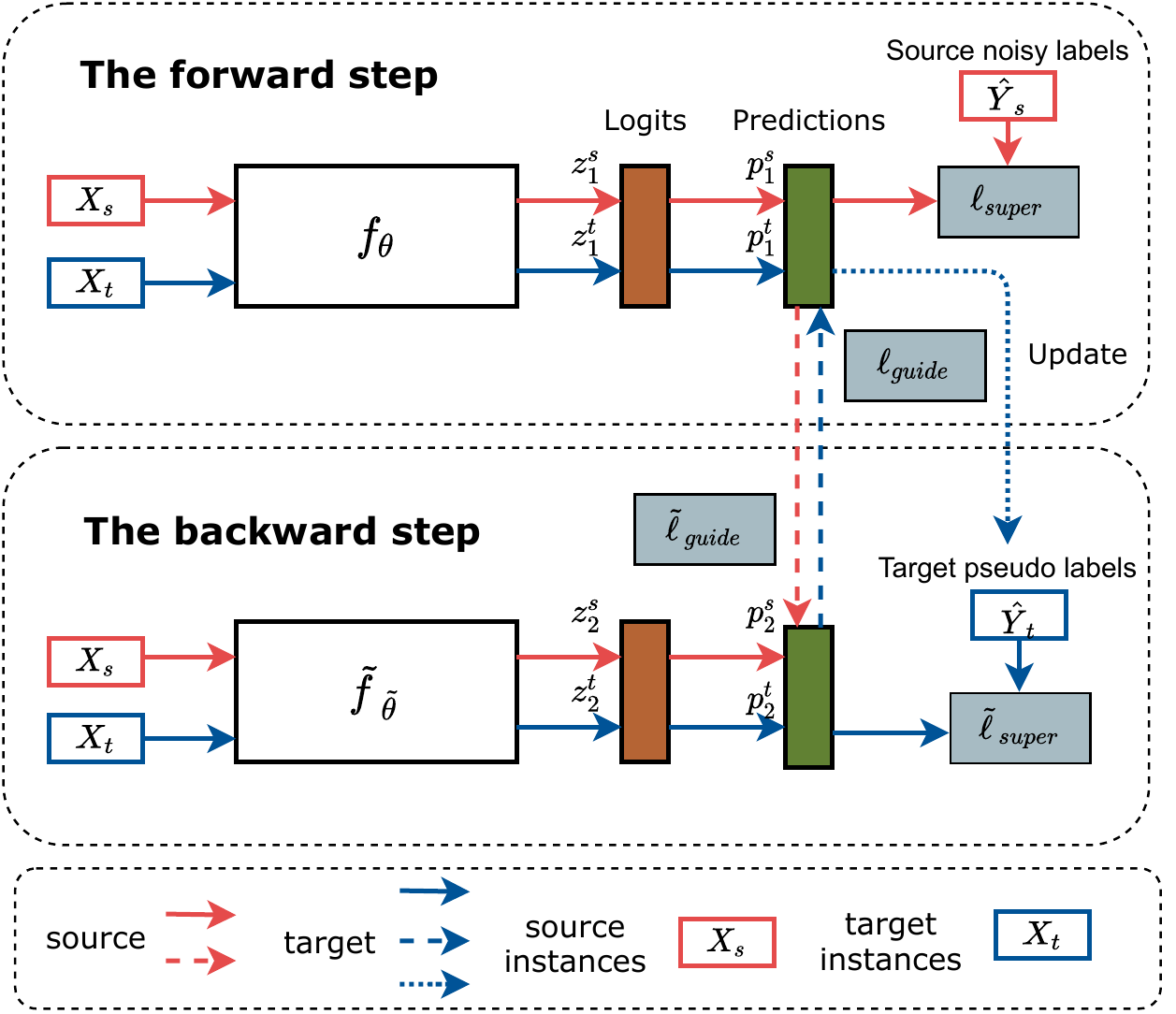}}
    \caption{GearNet schematic. The forward step and the backward step are conducted iteratively. Each model is trained with a supervised learning loss on one domain, and a symmetric Kullback Leibler divergence loss to mimic the predictions of its dual model on the other domain, where $z_{*}^{*}$ is the logit coming from the last layer of the corresponding model, and $\boldsymbol{p}_{*}^{*}$ is the probability of classes calculated by the softmax function on $z_{*}^{*}$. Every time when the \emph{forward step} stops, the pseudo labels of the target domain should be updated.}
    \label{fig:gearnet}
\end{figure}

\section{The Proposed GearNet}
% \subsection{Problem formulation}
% In this subsection, we introduce weakly 
% This paper proposed a method to address the issue of weakly-supervised domain adaptation in which labels from the source domain are partially corrupted, while the target domain suffers from the absence of labels. It is meaningful to handle this problem in the real world because highly-qualified source domains are difficult to obtain. 
% In this subsection, we first introduce the definition of weakly-supervised domain adaptation problem, the generation of two types of noisy labels and the notations used in this paper.
\noindent\textbf{Problem statement.}\quad Throughout this paper, we consider the classification task under the setting of WSDA.
% , where $\mathcal{X}$ is the input space and $\mathcal{Y}$ is the output space. 
We assume that we have a source domain with noisy labels corrupted from ground-truth labels $\hat{S}=\{(x_i^s, \hat{y}_i^s)\}_{i=1}^{n_s}$ and an unlabeled target domain $T=\{(x_i^t)\}_{i=1}^{n_t}$, where $n_s$ and $n_t$ denote the number of instances from the source domain and the target domain respectively, and $\hat{y}_i^s$ denotes the noisy (corrupted) label. 
% When label noise exists, we can only observe the corrupted source domain $\hat{D}_s$ from the corrupted distribution $p_s(x^s,\hat{y}^s)$, where $\hat{y}^s$ denotes the noisy (corrupted) label. Then, the source domain data with noisy labels are independently drawn from $\hat{D}_s$ while the unlabeled target domain data are drawn i.i.d. from $D_t^X$, where $D_t^X$ follows the marginal distribution of $p_t(x^t,y^t)$ over $X$.
% $$
%     \hat{S}={\{(x_i^s, \hat{y}_i^s)\}^n_{i=1}} \stackrel{\text { i.i.d. }}{\sim} (\hat{D}_s)^n;\quad T={\{(x_i^t)\}^{n^{'}}_{i=1}} \stackrel{\text { i.i.d. }}{\sim} (D_t^X)^{n^{'}}
% $$
% Due to the existence of label noise, the ground-truth label of the source domain will flip to another label with the conditional probability of $p(\hat{y}_i^s|y_i^s)$, where $\hat{y}_i^s$ is the corrupted labels. Therefore, in the problem of WSDA, we can actually observe data from the source domain as $\hat{S}={(x_i^s, \hat{y}_i^s)^n_{i=1}}$. 
Our goal is to train a classifier $f_{\theta}: X \rightarrow Y$ based on $\hat{S}$ and $T$ to accurately annotate samples from the target domain.
% with low target risk:
% $$
%     \mathcal{R}_{D_t}(f_{\theta})=\mathbb{E}_{(\mathbf{x}, \mathbf{y}) \sim D_t}\left[\left(f_{\theta}(
%     \mathbf{x})\right)\neq \mathbf{y}\right],
% $$
% where $D_t$ is the invisible labeled target domain.
% Given $D_s={(x_i^s, \tilde{y_i}^s)}_{i=1}^{n_s}$ drawn from the source domain following the distribution $\tilde{p_s}$, and a set of data $D_t={(x_i^t)}_{i=1}^{n_t}$ drawn from the target domain  following the distribution $p_t$, where $\tilde{p_s} \neq p_t$, $\tilde{p_s}$ is the distribution regarding noisy source data, and $\tilde{y_i}^s$ is the corrupted source labels, our goal is to train a classifier on $D_s$ and $D_t$ to accurately annotate data drawn from the target domain. 

As label noise will degenerate both the domain adaptation process \cite{yu2020label} and the classification process \cite{zhang2016understanding}, existing methods \cite{shu2019transferable, liu2019butterfly, yu2020label} for WSDA focus on emphasizing useful samples by utilizing the supervision information from the source domain to reduce the negative impact of label noise. Considering the source domain could provide useful information to the target domain as mentioned above, we claim the target domain could also contain valuable information that is beneficial to the learning on the source domain. Therefore, inspired by the mutual learning \cite{zhang2018deep} and dual learning \cite{luo2019dual}, we address the issue of WSDA by exploring the bilateral relationship that the two domains offer useful information to each other to handle domain shifts and label noise.

The intuition for exploring bilateral relationships in WSDA is briefly explained as follows. Similar to learning with label noise, existing WSDA methods would encounter the error accumulation issue: the error that comes from the biased selection of training instances in the previous iterations would be directly learnt again in the following training \cite{han2018co}. 
% When the error is transferred to the target domain, the biased prediction for the target domain instances is emerged. 
In WSDA, the accumulated error from learning with source domain examples would be amplified, causing a significant increase in the target domain error \cite{liu2019butterfly, han2020towards}.
% To handle this problem, we propose to leverage bilateral relationships that train two neural networks with opposite transfer directions. 
Co-teaching \cite{han2018co} and Butterfly \cite{liu2019butterfly} alleviate this issue by training two networks with different initialization to exchange the biased selections with each other.
In this work, our method introduce additional model diversity derived from the distinct supervision information by training two networks with opposite transfer directions.
In this manner, the accumulated error would be further attenuated during the training stage.

\begin{algorithm}[t]
  \SetKwInOut{Input}{input}
  \SetKwInOut{Output}{output}
\caption{\textcolor{black}{GearNet's Learning Strategy}}
\SetAlgoLined
\setstretch{1.1}

\label{GearNet}
\Input{Source dataset with noisy labels $\hat{S}$, target dataset with pseudo labels $\hat{T}$, max steps $M$, max epochs $N$, learning rate $\eta$, the pretrained basic model $f_{\theta}$ and its dual model $\tilde{f}_{\tilde{\theta}}$}
\Output{$f_{\theta}$ and $\tilde{f}_{\tilde{\theta}}$}

  \BlankLine
\For{$t=0$ $\mathrm{to}$ $M$}{
 Shuffle: $\hat{S}$ and $\hat{T}$\;
 Initialize: $\tilde{f}_{\tilde{\theta}}$ \tcp*{Start the backward step}
\For{$i=0$ $\mathrm{to}$ $N$}{ 
Fetch: $\{x_i^t, \hat{y}_i^t\}_{i=1}^{m_t}$ from $\hat{T}$, $\{x_i^s\}_{i=1}^{m_s}$ from $\hat{S}$\;
Calculate: $\tilde{\ell}_{\mathrm{super}}=\frac{1}{m_t}\sum_{i=1}^{m_t}\ell(\hat{y}_i^t,\tilde{f}(x_i^t,\tilde{\theta})$)\;
Forward: $\boldsymbol{p}_1^s = f(x_i^s, \theta)$, $\boldsymbol{p}_2^s = \tilde{f}(x_i^s, \tilde{\theta})$\;
Calculate: $\tilde{\ell}_{guide}$ by (\ref{agree_2}) using $\boldsymbol{p}_1^s$ and $\boldsymbol{p}_2^s$\;
Obtain: $\tilde{\ell}_{total}$ by (\ref{total}) \;
Update: $\tilde{\theta}=\tilde{\theta}-\eta \Delta \tilde{\ell}_{total} $\;
}

Initialize: $f_\theta$\ \tcp*{Start the forward step}
\For{$i=0$ $\mathrm{to}$ $N$}{
Fetch: $\{x_i^s, \hat{y}_i^s\}_{i=1}^{m_s}$ from $\hat{S}$, $\{x_i^t\}_{i=1}^{m_t}$ from $\hat{T}$\;
Calculate: $\ell_{\mathrm{super}}=\frac{1}{m_s}\sum_{i=1}^{m_s}\ell(\hat{y}_i^s,f(x_i^s,\theta)$)\;
Forward: $\boldsymbol{p}_1^t= f(x_i^t, \theta)$, $\boldsymbol{p}_2^t = \tilde{f}(x_i^t, \tilde{\theta})$\;
Calculate: $\ell_{guide}$ by (\ref{agree_1}) using $\boldsymbol{p}_1^t$ and $\boldsymbol{p}_2^t$\;
Obtain: $\ell_{total}$ by (\ref{total})\;
Update: $\theta=\theta-\eta \Delta \ell_{total}$\;}
Update: $\{\hat{y}_i^t\}_{i=1}^{n_t}$ by $f_{\theta}$\;}
\end{algorithm}

\noindent\textbf{Algorithm design.}\quad Inspired by the above motivation, we propose a universal paradigm called "GearNet" that can be employed to various backbone methods (i.e., existing WSDA methods). Before the introduction, we need to clarify at first that we omit the technical details of those backbone methods to simplify the introduction of GearNet, and assume that we build GearNet on the top of a basic backbone model that is composed of a feed-forward neural network. With this basic model, we can introduce GearNet in a more convenient way. After the introduction, we will further introduce how to employ GearNet to different backbone methods.

% composed of a \emph{forward pass} and a \emph{backward pass} for every mini-batch, where the forward pass returns loss values by feeding data of this mini-batch though the model, and the backward pass returns the gradient of the loss values by propagating the loss. With this basic algorithm, we can introduce GearNet in a more convenient way. After the introduction, we can further introduce how to employ GearNet to different backbone methods.

Our GearNet compromises two basic models: $f_{\theta}$ and $\tilde{f}_{\tilde{\theta}}$. Our model learning strategy includes three parts: the \emph{pretrained process} aiming to annotate the target domain data by $f_{\theta}$ , the \emph{forward step} aiming to transfer knowledge from the source domain to the target domain by $f_{\theta}$, and the \emph{backward step} aiming to transfer knowledge from the target domain to the source domain by $\tilde{f}_{\tilde{\theta}}$. The forward step and the backward step are iteratively conducted until the whole training process ends.
% we propose the universal paradigm 

% which is composed of three parts: \emph{the pretrained process} aiming to annotate the target domain data, \emph{the forward step} aiming to transfer knowledge from the source domain to the target domain, and \emph{the backward step} aiming to transfer knowledge from the target domain to the source domain. A prototype of GearNet is given in Algorithm \ref{GearNet} except the first part.
% As this method is an universal algorithm that can be used with other backbone approaches, the difference for various backbone methods depends on the basic models $f_\theta$ and $\tilde{f}_{\tilde{\theta}}$,  and the inner loss term. The whole algorithm is composed of three parts: \emph{the pretrained process} that aims to annotate the target domain data, \emph{the forward step} that aims to transfer knowledge from the source domain to the target domain, and \emph{the backward step} that aims to transfer knowledge from the target domain to the source domain. A prototype of GearNet is given in Algorithm \ref{GearNet} except the first part.

In the \emph{pretrained process}, we train $f_{\theta}$ on the source domain data with noisy labels $\hat{D}_s$ (i.e., Eq. (\ref{basic_loss})), and then use the model to generate the hard pseudo labels for the target domain instances (i.e., Eq. (\ref{pseudo})).
% , where the basic algorithm could be expressed as Eq. (\ref{basic_loss}) and the pseudo label generation is as Eq. (\ref{pseudo}):
\begin{equation}
\label{basic_loss}
 \theta = \underset{\theta}{\operatorname{argmin}}\frac{1}{m_s}\sum\nolimits_{i=1}^{m_s}\ell(\hat{y}_i^s,f(x_i^s,\theta)),
\end{equation}
\begin{equation}
\label{pseudo}
 \hat{y}_i^t=\underset{c}{\operatorname{argmax}}f_{c}(x_i^t, \theta), \forall i=1,2,..., n_t,
\end{equation}
where $c$ denotes the $c$'{th} label, and $f_{c}$ is the network output for the $c$'{th} label.

Then the forward step and the backward step can be conducted in an opposite manner. In detail, both the \emph{forward} and the \emph{backward step} train their models with two losses: a conventional supervised learning loss (i.e., $\ell_{super}$ or $\tilde{\ell}_{super}$) on one domain and a mimicry loss that aligns predictions of the two models (i.e., $\ell_{guide}$ or $\tilde{\ell}_{guide}$) on the other domain. So their overall losses could be expressed as follows:
\begin{equation}
\label{total}
    \ell_{\text {total}}=\ell_{\text {super}}+\beta \ell_{\text {guide}};\quad \tilde{\ell}_{\text {total}}=\tilde{\ell}_{\text {super}}+\beta \tilde{\ell}_{\text {guide}},
\end{equation}
where $\beta$ is the trade-off hyperparameter, and its value is set as 0.1 in general. Besides, $\ell_{\text {total}}$ is for training $f_{\theta}$ during the forward step, while $\tilde{\ell}_{\text {total}}$ is for training $\tilde{f}_{\tilde{\theta}}$ during the backward step.

The supervised learning loss of the forward step is based on the noisy source domain:
\begin{equation}
\label{f1_l1}
    \ell_{\mathrm{super}}=\frac{1}{m_s}\sum\nolimits_{i=1}^{m_s}\ell(\hat{y}_i^s,f(x_i^s,\theta)),
\end{equation}
while that of the backward step is based on the pseudo-labeled target domain:
\begin{equation}
\label{f2_l1}
    \tilde{\ell}_{\mathrm{super}}=\frac{1}{m_t}\sum\nolimits_{i=1}^{m_t}\ell(\hat{y}_i^t,\tilde{f}(x_i^t,\tilde{\theta})).
\end{equation}
 
Although the model can perform well on the domain with supervision information due to the optimization of the supervised learning loss, there would be a significant accuracy drop on the test data from the other domain because of domain shifts. To generalize the model for better performance on the other domain, we introduce a consistency regularization, the symmetric Kullback-Leibler (KL) divergence loss (i.e., $\ell_{guide}$ or $\tilde{\ell}_{guide}$), which can enforce the model to mimic the predictions of its dual model for every sample from the other domain. So for the \emph{forward step}, the loss is based on the target domain:
\begin{equation}
\label{agree_1}
    \ell_{\text {guide}}=D_{\mathrm{KL}}\left(\boldsymbol{p}_{1}^t \| \boldsymbol{p}_{2}^t\right)+D_{\mathrm{KL}}\left(\boldsymbol{p}_{2}^t \| \boldsymbol{p}_{1}^t\right),
\end{equation}
% where 
% \begin{equation}
%     D_{\mathrm{KL}}\left(\boldsymbol{p}_{1}^t \| \boldsymbol{p}_{2}^t\right)=\sum_{i=1}^{m_t} \sum_{c=1}^{C} p_{1}^{c}\left(x_{i}^{t}\right) \log \frac{p_{1}^{c}\left(x_{i}^{t}\right)}{p_{2}^{c}\left(x_{i}^{t}\right)},
% \end{equation}
% \begin{equation}
%     D_{\mathrm{KL}}\left(\boldsymbol{p}_{2}^t\| \boldsymbol{p}_{1}^t\right)=\sum_{i=1}^{m_t} \sum_{c=1}^{C} p_{2}^{c}\left(x_{i}^{t}\right) \log \frac{p_{2}^{c}\left(x_{i}^{t}\right)}{p_{1}^{t}\left(x_{i}^{t}\right)}.
% \end{equation}

For the \emph{backward step}, the loss is calculated by the data from the source domain:
\begin{equation}
\label{agree_2}
    \tilde{\ell}_{\text {guide}}=D_{\mathrm{KL}}\left(\boldsymbol{p}_{1}^s \| \boldsymbol{p}_{2}^s\right)+D_{\mathrm{KL}}\left(\boldsymbol{p}_{2}^s \| \boldsymbol{p}_{1}^s\right).
\end{equation}
% where
% \begin{equation}
%     D_{\mathrm{KL}}\left(\boldsymbol{p}_{1}^s \| \boldsymbol{p}_{2}^s\right)=\sum_{i=1}^{m_s} \sum_{c=1}^{C} p_{1}^{c}\left(x_{i}^{s}\right) \log \frac{p_{1}^{c}\left(x_{i}^{s}\right)}{p_{2}^{c}\left(x_{i}^{s}\right)},
% \end{equation}
% \begin{equation}
%     D_{\mathrm{KL}}\left(\boldsymbol{p}_{2}^s\| \boldsymbol{p}_{1}^s\right)=\sum_{i=1}^{m_s} \sum_{c=1}^{C} p_{2}^{c}\left(x_{i}^{s}\right) \log \frac{p_{2}^{c}\left(x_{i}^{s}\right)}{p_{1}^{c}\left(x_{i}^{s}\right)}.
% \end{equation}

$D_{\mathrm{KL}}$ denotes the Kullback-Leibler (KL) divergence that measures the probability difference \cite{kullback1951information}:
\begin{equation}
\label{kl}
    D_{\mathrm{KL}}\left(p\|q\right)= \sum_{i=1}^n p(x_i)log\frac{p(x_i)}{q(x_i)},
\end{equation}
where $p$ and $q$ denote the probability to be measured for the probability difference, and $n$ is the number of samples.  

In the \emph{forward step}, the consistency regularization is obtained by inputting $\boldsymbol{p}_1^t$ and $\boldsymbol{p}_2^t$ into Eq. (\ref{kl}), where $\boldsymbol{p}_1^t$ and $\boldsymbol{p}_2^t$ denotes class label distributions which are outputs from $f_{\theta}$ and $\tilde{f}_{\tilde{\theta}}$, respectively. In the \emph{backward step}, the consistency regularization is calculated by inputting $\boldsymbol{p}_1^s$ and $\boldsymbol{p}_2^s$, where $\boldsymbol{p}_1^s$ and $\boldsymbol{p}_2^s$ denote the class label distributions which are outputs from $f_{\theta}$ and $\tilde{f}_{\tilde{\theta}}$, respectively.
% Note that $\boldsymbol{p}_1^s$ and $\boldsymbol{p}_1^t$ are the probability of all classes obtained by feeding the source and the target domain mini-batch into $f_{\theta}$, respectively, while $\boldsymbol{p}_2^s$ and $\boldsymbol{p}_2^t$ denote the same meaning from $\tilde{f}_{\tilde{\theta}}$. In the \emph{forward step}, the symmetric KL Divergence is obtained by feeding the target domain data, while in the \emph{backward step}, it is calculated by feeding the source domain data.
% However, we use the symmetric version as Eq. (\ref{agree_1}) and Eq. (\ref{agree_2}) to measure the probability discrepancy for both models. On one hand, the values of KL divergence are different for the two models, since the it is asymmetric \cite{zhang2018deep}; on the other hand, we are not confident about the default distribution provided by the dual model. 
 Every factor in those probability metrics is computed by the softmax function based on the corresponding logits. For example, the probability of class $c$ for the sample $x_i^s$ from $f_{\theta}$ (i.e., $p_1^c(x_i^s)$) is calculated as:
\begin{equation}
    p_1^c(x_i^s)=\frac{exp(z_1^{c})}{\sum_{c^{'}=1}^{C}exp(z_1^{c^{'}})},
\end{equation}
where $z_1^{c}$ denotes the logit of class $c$ from $f_{\theta}$ for $x_i^s$.

% Different from the criteria to reduce the discrepancy between two different domains in the problem of domain adaptation, the consistency regularization is for matching the predictions between two models from the same domain.  
% Inspired from mutual learning \cite{}

\noindent\textbf{Optimisation of GearNet.}\quad After the \emph{pretrained process} to provide pseudo labels to the target domain, the whole algorithm is run as Algorithm \ref{GearNet} and Figure \ref{fig:gearnet}. We first conduct the \emph{backward step} by computing the total loss $\tilde{\ell}_{total}$ as Eq. (\ref{total}) for training $\tilde{f}_{\tilde{\theta}}$, where the second loss $\tilde{\ell}_{guide}$ is between $\tilde{f}_{\tilde{\theta}}$ and the pre-trained $f_{\theta}$ on the source domain. Then we can continue to update the parameters of $f_{\theta}$ in the \emph{forward step} by the total loss $\ell_{total}$ also as Eq. (\ref{total}), where the mimicry loss $\ell_{guide}$ is between the initialized $f_{\theta}$ and $\tilde{f}_{\tilde{\theta}}$ trained during the \emph{backward step} on the target domain, after which we update the pseudo labels of the target domain by the trained $f_{\theta}$. We repeat the \emph{forward} and the \emph{backward steps} until this algorithm stops. It is worthy to note that the training model should be initialized before its training process to avoid overfitting to the noisy samples.

 \begin{table*}[h]\tiny
\small
\begin{center}
\setlength{\tabcolsep}{4pt} % Default value: 6pt
\renewcommand{\arraystretch}{1} % Default value: 1
\resizebox{!}{!}{ %
\begin{tabular}{l|ccccccc}
\toprule
 Tasks &$A\rightarrow W$  & $A\rightarrow D$  & $W\rightarrow A$  & $W\rightarrow D$  & $D\rightarrow A$  & $D\rightarrow W$  & Average \tabularnewline
\midrule
\parbox[t]{15mm}{\multirow{1}{*}{\rotatebox[origin=c]{0}{Standard}}}
 & $46.61 \pm 0.32$ & $51.46 \pm 0.53$ & $44.21 \pm 0.12$ & $73.13 \pm 0.26$ & $43.43 \pm 0.32$ & $65.63 \pm 0.41$ & $54.06 \pm 0.33 $\tabularnewline
% \midrule

\parbox[t]{15mm}{\multirow{1}{*}{\rotatebox[origin=c]{0}{Co-teaching}}}
  &$49.87 \pm 1.42$  & $55.00 \pm 0.89$ & $42.18 \pm 0.71$ & $75.63 \pm 0.85$ & $44.85 \pm 1.01$ & $64.06 \pm 2.01$ & $55.98 \pm 1.15$\tabularnewline
% \midrule

 \parbox[t]{15mm}{\multirow{1}{*}{\rotatebox[origin=c]{0}{JoCoR}}}

 & $50.53 \pm 1.67$  & $55.42 \pm 2.33$ & $47.19 \pm 1.71$ & $74.79 \pm 1.28$ & $44.50 \pm 1.01$    & $61.72 \pm 0.98$  & $55.69 \pm 1.50$ \tabularnewline

% \midrule

\parbox[t]{15mm}{\multirow{1}{*}{\rotatebox[origin=c]{0}{DAN}}}
 & $54.39 \pm 2.11$   & $54.79 \pm 1.32$ & $36.65 \pm 2.62$ & $67.08 \pm 1.79$ & $35.09 \pm 1.58$  & $60.94 \pm 2.06$ & $51.32 \pm 1.91$\tabularnewline

% \midrule
\parbox[t]{15mm}{\multirow{1}{*}{\rotatebox[origin=c]{0}{DANN}}}
 &$50.91 \pm 1.88$   & $54.17 \pm 0.87$ & $44.57 \pm 0.74$ & $74.79 \pm 1.08$ & $45.35 \pm 1.41$  & $67.58 \pm 0.48$ & $56.23 \pm 1.08$\tabularnewline

% \midrule
\parbox[t]{15mm}{\multirow{1}{*}{\rotatebox[origin=c]{0}{TCL}}}
 & $56.46 \pm 0.67$   & $63.13 \pm 1.14$ & $45.31 \pm 0.31$ & $76.87 \pm 0.85$ & $44.78 \pm 0.60$  & $71.22 \pm 0.58$ & $59.63 \pm 0.69$\tabularnewline

% \midrule
% \parbox[t]{15mm}{\multirow{1}{*}{\rotatebox[origin=c]{0}{RDA}}}
% & $62.76$   & $n$ & $44.24$ & $N$ & $N$  & $N$ & $N$\tabularnewline
% \hline

% \parbox[t]{15mm}{\multirow{1}{*}{\rotatebox[origin=c]{0}{Butterfly}}}
%  &$N$   & $N$ & $N$ & $N$ & $N$  & $N$ & $N$\tabularnewline

\midrule
% \midrule
\parbox[t]{15mm}{\multirow{1}{*}{\rotatebox[origin=c]{0}{$\text{GearNet}_{\text{Co-teaching}}$}}}
& $53.12 \pm 1.88$   & $58.12 \pm 1.11$ & $44.49 \pm 0.57$ & $76.87 \pm 1.94$ & \textbf{49.28$\pm$1.37}  & $69.14 \pm 1.81$ & $56.75 \pm 1.44$\tabularnewline
  
\parbox[t]{15mm}{\multirow{1}{*}{\rotatebox[origin=c]{0}{$\text{GearNet}_{\text{DANN}}$}}}
 &\textbf{60.68 $\pm$ 0.26}   & $63.54 \pm 1.03$ & $47.19 \pm 0.96$ & $76.88 \pm 1.68$ & 47.90 $\pm$ 0.66  & $72.39 \pm 0.79$ & $61.43 \pm 0.90$\tabularnewline
 
 \parbox[t]{28mm}{\multirow{1}{*}{\rotatebox[origin=c]{0}{$\text{GearNet}_{\text{TCL}}$}}}
  &$58.84 \pm 0.57$   & \textbf{65.63 $\pm$ 0.93} & \textbf{48.37 $\pm$ 1.04} &  \textbf{78.54 $\pm$ 0.75}   & $47.80 \pm 0.58$  & \textbf{73.44 $\pm$ 0.56}& \textbf{62.10 $\pm$ 0.74} \tabularnewline
 \bottomrule 
% \parbox[t]{15mm}{\multirow{1}{*}{\rotatebox[origin=c]{0}{G+Butterfly}}}
%   &$N$   & $N$ & $N$ & $N$ & $N$  & $N$ & $N$\tabularnewline
% \hline 
% \hline 
\end{tabular}}
\par\end{center}
\caption{Target accuracy (\%) on Office-31 datasets with Unif-20\% noise. Bold numbers are superior results}
 \label{unif2}
\end{table*}

\noindent\textbf{Realizations of GearNet.}\quad In this subsection, we incorporate three backbone methods with GearNet as examples. They are Co-teaching \cite{han2018co} that belongs to the approach to improve model robustness against label noise, DANN \cite{ganin2016domain} that belongs to the domain adaptation approach and TCL \cite{shu2019transferable} that belongs to the WSDA approach, respectively. All the three algorithms are representative approaches, which could spotlight the universal capability of GearNet.

First of all, we also need to initialize two models $f_{\theta}$ and $\tilde{f}_{\tilde{\theta}}$ with the backbone algorithm. Although these backbone methods have different learning strategies, we generally express their loss functions as follows to highlight the structure of GearNet:
\begin{equation}
    \ell_{bone}=\mathbb{E}_{p^s(x^s,\hat{y^s}),p^t(x^t)}(\ell(x^s, \hat{y^s}, x^t;f_{\theta})),
\end{equation}
\begin{equation}
    \tilde{\ell}_{bone}=\mathbb{E}_{p^t(x^t,\hat{y^t}),p^s(x^s)}(\ell(x^t, \hat{y^t}, x^s;\tilde{f}_{\tilde{\theta}})),
\end{equation}
where $\ell_{bone}$ denotes the loss of the backbone method for training $f_{\theta}$, while $\tilde{\ell}_{bone}$ denotes the same meaning for training $\tilde{f}_{\tilde{\theta}}$. Besides, $p^s(*)$ and $p^t(*)$ denote the distribution from the source domain, and the target domain, respectively.
% $p_s(x^s,\hat{y^s})$ and $p_t(x^t,\hat{y^t})$ denote the joint distribution of the noisy source domain and the pseudo-labeled target domain, respectively, while $p_s(x^s)$ and $p_t(x^t)$ denote the marginal distribution of the source and the target domains.

The two losses above take the place of $\ell_{super}$ and $\tilde{\ell}_{super}$ in Eq. (\ref{total}) when we chose those backbone methods. They represent different meanings under various backbone algorithms. For the Co-teaching backbone method, they reduce the impact of noise by cross-updating two peer networks. As for DANN, they decrease the domain discrepancy using a domain discriminator in an adversarial manner. For TCL, they address both the label noise problem and the domain shift problem by selecting noiseless and transferable samples from the source domain to train the DANN-shape model. 

To obtain $\ell_{guide}$ and $\tilde{\ell}_{guide}$, the training model should align its predictions with corresponding class posteriors of its dual model. Note that only classification predictions need to be aligned. Besides, for multi-classifier models, like Co-teaching, these losses should be computed for every classifier with the dual model, so that all of them can have the professional guidance on the domain that they are not good at. 
% For the loss $loss_{guide}$, it should be calculated between corresponding classifiers of the two models. For example, when equipping DANN with GearNet, we feed the feature vectors of the relevant domain into the two DANN models and keep the outputs from the classification layer as $p_1(x_i^s)$ and $p_2(x_i^s)$. And then we can calculate the symmetric KL divergence loss like Eq. (\ref{agree}). For the multi-classifier learning algorithm, like Co-teaching which contains two neural networks, $loss_{guide}$ should be computed for every classifier, so that all classifiers can obtain the professional guidance.

% \noindent\textbf{Relations to Coteaching.} In learning from noisy labels, the error flow comes from the biased selection of training instances in the first iteration. In existing WSDA methods that only exploit unidirectional relationships, the error from one network trained on source domain will be directly transferred back to itself in the next iteration, and the error should be increasingly accumulated. Co-teaching and Butterfly reduce this error by introducing model uncertainty, where they train two networks with different initialization to exchange the biased selections with each other. In our method, training two networks with opposite transfer directions introduce additional diversity derived from the distinct supervised information. In this manner, the accumulated error would be largely attenuated during the training stage.

\noindent\textbf{Relation to CycleGAN.}\quad CyCADA \cite{hoffman2018cycada} and Bi-Directional Generation domain adaptation model (BGD) \cite{yang2020bi} also propose the idea that transfers knowledge from the target domain to the the source domain, which are inspired by CycleGAN \cite{zhu2017unpaired}. They transfer the instances of the target domain to that of the source domain by a feature generator, which aim is to obtain a feature space that is close to the source domain. However, there are fundamental differences between them and GearNet. (i) CyCADA and BGD obtain a feature generator that can convert the target-style feature to the source-style feature, but GearNet trains a model that leverages the target domain to predict labels of the source domain. (ii) The reason why CyCADA and BGD transfer the feature style is to predict the labels of the target domain using the classifier trained by the source domain. But GearNet aims to exploit information from the target domain which could enhance both the domain adaptation process and the noise reduction process for the source domain. (iii) CyCADA and BGD address the issue of unsupervised domain adaptation, but GearNet handles the issue of weakly-supervised domain adaptation. 

\noindent\textbf{Relation to multi-task learning.}\quad In the problem of multi-task learning, there is also an idea that a noisy task can reduce its noise under the assistance of another noisy task \cite{wu2020understanding}. However, the crucial difference between multi-task learning and GearNet is that multi-task learning aims at good performance for all the tasks, but GearNet only needs to achieve good performance for the target domain. In addition, the above idea for multi-task learning proposes that more noisy tasks can reduce the impact of noise and get better performance by up weighting less noisy tasks, but GearNet proposes that both  the target domain and the source domain contain useful knowledge for each other. 

\section{Experiments}
\begin{figure}
\begin{subfigure}{.48\textwidth}
\centering
\subfloat[]{\includegraphics[width=.47\linewidth]{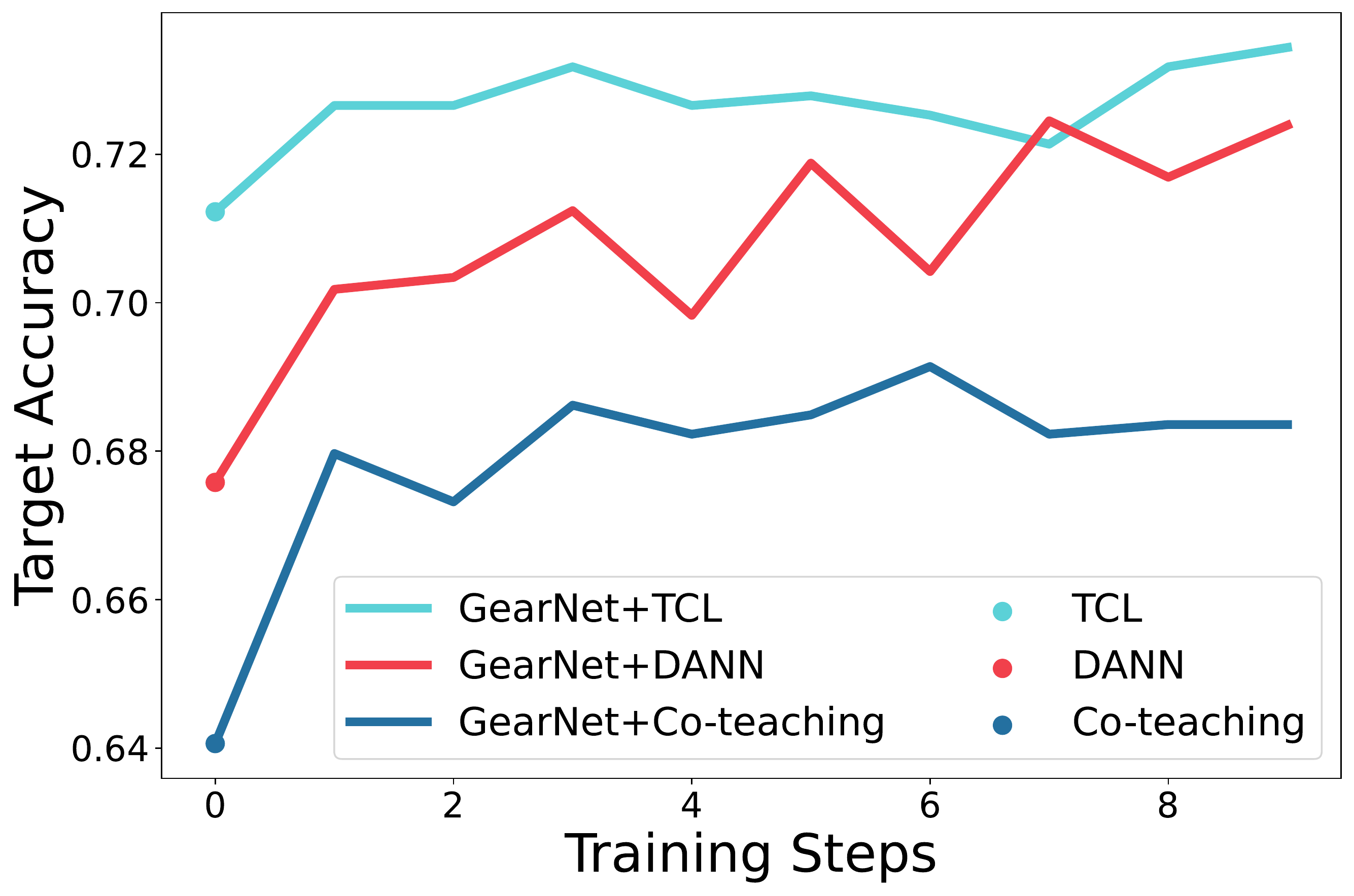}}\quad
\subfloat[]{\includegraphics[width=.47\linewidth]{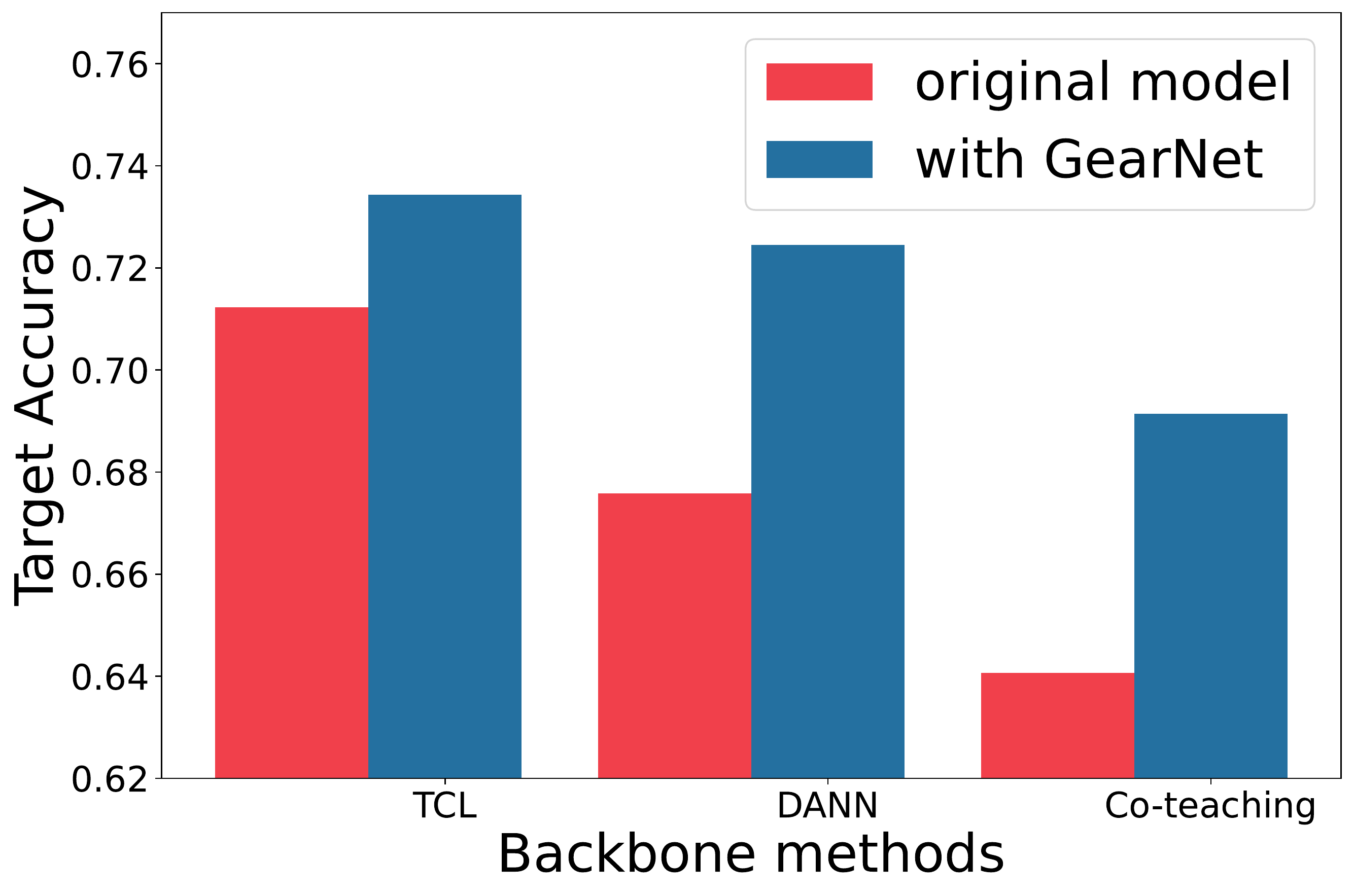}}
\end{subfigure}
\caption{Universal capability of GearNet on $D \rightarrow W$ with Unif-20\% noise under WSDA (TCL), de-noise (Co-teaching) and UDA (DANN) backbone methods. (a) Trend of target accuracy across training steps. (b) Best target accuracy across training steps}
\label{fig:20}
% \begin{minipage}{.48\textwidth}
% \centering
% \subfloat[]{\includegraphics[width=.4\linewidth]{fig3 (1).pdf}}\quad
% \subfloat[]{\includegraphics[width=.4\linewidth]{fig4 (1).pdf}}
% \caption{(a):Target accuracy during training on $D \rightarrow W$ under 40\% uniform noise rate. (b):Best target accuracy of three backbone methods and GearNet on $D \rightarrow W$ under 40\% uniform noise rate.}
% \label{fig:40}
% \end{minipage}
\end{figure}
 \begin{table*}[h]
\small
\begin{center}
\setlength{\tabcolsep}{4pt} % Default value: 6pt
\renewcommand{\arraystretch}{1} % Default value: 1
\resizebox{!}{!}{ %
\begin{tabular}{l|ccccccc}
\toprule
 Tasks &$A\rightarrow W$  & $A\rightarrow D$  & $W\rightarrow A$  & $W\rightarrow D$  & $D\rightarrow A$  & $D\rightarrow W$  & $\text{Average}$ \tabularnewline
\midrule
\parbox[t]{15mm}{\multirow{1}{*}{\rotatebox[origin=c]{0}{Standard}}}
 & $34.37 \pm 0.78$ & $36.45 \pm 0.43$ & $29.26 \pm 0.52$ & $54.79 \pm 0.72$ & $30.79 \pm 0.95$ & $48.31 \pm 0.69$ & $40.00 \pm 0.68 $\tabularnewline
% \midrule 

\parbox[t]{15mm}{\multirow{1}{*}{\rotatebox[origin=c]{0}{Co-teaching}}}
  &$36.20 \pm 2.45$  & $41.04 \pm 1.42$ & $31.11 \pm 1.64$ & $53.13 \pm 1.91$ & $23.08 \pm 1.63$ & $38.93 \pm 2.63$ & $37.25 \pm 1.95$\tabularnewline
% \midrule

 \parbox[t]{15mm}{\multirow{1}{*}{\rotatebox[origin=c]{0}{JoCoR}}}

 &$37.11 \pm 1.27$   & $42.29 \pm 2.24$ & $28.98 \pm 1.78$ & $47.50 \pm 1.89$ & $23.40 \pm 2.09$  & $36.85 \pm 2.74$ & $36.02 \pm 2.00$ \tabularnewline

% \midrule

\parbox[t]{15mm}{\multirow{1}{*}{\rotatebox[origin=c]{0}{DAN}}}
 & $34.24 \pm 1.73$   & $35.83 \pm 2.42$ & $23.97 \pm 2.89$ & $47.71 \pm 1.99$ & $24.96 \pm 2.07$  & $41.02 \pm 1.79$ & $34.62 \pm 2.15$\tabularnewline

% \midrule
\parbox[t]{15mm}{\multirow{1}{*}{\rotatebox[origin=c]{0}{DANN}}}
 &$36.20 \pm 1.62$   & $40.20 \pm 1.27$ & $30.22 \pm 1.53$ & $54.38 \pm 1.59$ & $32.71 \pm 2.06$  & $49.74 \pm 1.82$ & $40.57 \pm 1.65$\tabularnewline

% \midrule
\parbox[t]{15mm}{\multirow{1}{*}{\rotatebox[origin=c]{0}{TCL}}}
 & $42.06 \pm 1.86$   &$46.04 \pm 2.68$ & $29.55 \pm 0.96$ & $54.38 \pm 1.74$ & $30.43 \pm 1.83$ & $49.09 \pm 2.62$   & $41.95 \pm 1.95$\tabularnewline

% \midrule
% \parbox[t]{15mm}{\multirow{1}{*}{\rotatebox[origin=c]{0}{RDA}}}
% &$43.09$   & $N$ & $N$ & $N$ & $N$  & $N$ & $N$\tabularnewline
% \hline

% \parbox[t]{15mm}{\multirow{1}{*}{\rotatebox[origin=c]{0}{Butterfly}}}
%  &$N$   & $N$ & $N$ & $N$ & $N$  & $N$ & $N$\tabularnewline

\midrule

% \midrule
\parbox[t]{15mm}{\multirow{1}{*}{\rotatebox[origin=c]{0}{$\text{GearNet}_{\text{Co-teaching}}$}}}
& $39.32 \pm 1.07$   & $43.33 \pm 0.97$ & \textbf{33.94 $\pm$ 1.07} & $56.04 \pm 1.31$ & $25.74 \pm 1.58$  & $38.41 \pm 1.30$ & $39.46 \pm 1.21$\tabularnewline

\parbox[t]{15mm}{\multirow{1}{*}{\rotatebox[origin=c]{0}{$\text{GearNet}_{\text{DANN}}$}}}
 &$43.48 \pm 1.45$   & \textbf{48.54 $\pm$ 1.72} & $30.45 \pm 1.37$ & \textbf{58.54 $\pm$ 1.86} & \textbf{35.51 $\pm$ 0.83}  & \textbf{54.17 $\pm$ 1.06} & \textbf{45.12 $\pm$ 1.38}\tabularnewline

\parbox[t]{28mm}{\multirow{1}{*}{\rotatebox[origin=c]{0}{$\text{GearNet}_{\text{TCL}}$}}}
  &\textbf{46.61 $\pm$ 0.89}   & $47.50 \pm 1.28$ & $30.39 \pm 2.01$ & $55.83 \pm 1.92$ & $28.91 \pm 1.47$  & $52.47 \pm 1.39$ & $43.62 \pm 1.49$ \tabularnewline
% \hline 
% \parbox[t]{15mm}{\multirow{1}{*}{\otatebox[origin=c]{0}{G+Butterfly}}}
%   &$N$   & $N$ & $N$ & $N$ & $N$  & $N$ & $N$\tabularnewline
\bottomrule
\end{tabular}}
\caption{Target accuracy (\%) on Office-31 datasets with Unif-40\% noise. Bold numbers are superior results.}
\label{unif4}
\par\end{center}
\end{table*}

 \begin{table*}[h]\tiny
 
\small
\begin{center}
\setlength{\tabcolsep}{4pt} % Default value: 6pt
\renewcommand{\arraystretch}{1} % Default value: 1
\resizebox{!}{!}{ %
\begin{tabular}{l|ccccccc}
\toprule
 Tasks &$A\rightarrow W$  & $A\rightarrow D$  & $W\rightarrow A$  & $W\rightarrow D$  & $D\rightarrow A$  & $D\rightarrow W$  & $\text{Average}$ \tabularnewline
\midrule
\parbox[t]{15mm}{\multirow{1}{*}{\rotatebox[origin=c]{0}{Standard}}}
 & $46.61 \pm 0.92$ & $46.67 \pm 1.83$ & $41.41 \pm 0.94$ & $69.17 \pm 1.03$ & $40.23 \pm 1.06$ & $64.19 \pm 0.74$ & $51.38 \pm 1.08 $\tabularnewline
% \midrule

\parbox[t]{15mm}{\multirow{1}{*}{\rotatebox[origin=c]{0}{Co-teaching}}}
  &$48.31 \pm 2.01$  & $50.21 \pm 1.37$ & $42.19 \pm 1.87$ & $69.17 \pm 2.40$ & $39.20 \pm 1.82$ & $58.59 \pm 2.72$ & $51.28 \pm 2.03$\tabularnewline
% \midrule

 \parbox[t]{15mm}{\multirow{1}{*}{\rotatebox[origin=c]{0}{JoCoR}}}
 &$49.74 \pm 0.98$   & $51.46 \pm 1.52$ & $41.94 \pm 1.83$ & $67.92 \pm 2.07$ & $37.71 \pm 1.68$  & $55.86 \pm 0.93$ & $50.77 \pm 1.50$ \tabularnewline
% \midrule

\parbox[t]{28mm}{\multirow{1}{*}{\rotatebox[origin=c]{0}{DAN}}}
 & $56.64 \pm 1.73$   & $53.54 \pm 2.04$ & $40.20 \pm 2.13$ & $70.21 \pm 1.57$ & $35.80 \pm 1.78$  & $64.84 \pm 1.84$ & $53.54 \pm 1.85$\tabularnewline
 
% \midrule
\parbox[t]{15mm}{\multirow{1}{*}{\rotatebox[origin=c]{0}{DANN}}}
 &$47.66 \pm 1.46$   & $50.63 \pm 1.76$ & $39.17 \pm 0.63$ & $69.58 \pm 1.05$ & $41.55 \pm 0.77$  & $65.49 \pm 0.86$ & $52.35 \pm 1.09$\tabularnewline

% \midrule
\parbox[t]{15mm}{\multirow{1}{*}{\rotatebox[origin=c]{0}{TCL}}}
 & $55.99 \pm 1.79$   & $61.04 \pm 1.53$ & $42.37 \pm 2.51$ & $72.92 \pm 0.62$ & $42.29 \pm 1.33$  & $70.96 \pm 2.52$ & $57.60 \pm 1.72$\tabularnewline

\midrule
% \parbox[t]{15mm}{\multirow{1}{*}{\rotatebox[origin=c]{0}{RDA}}}
% &$59.64$   & $N$ & $N$ & $N$ & $N$  & $N$ & $N$\tabularnewline
% \hline

% \parbox[t]{15mm}{\multirow{1}{*}{\rotatebox[origin=c]{0}{Butterfly}}}
% &$N$   & $N$ & $N$ & $N$ & $N$  & $N$ & $N$\tabularnewline

% \midrule

% \midrule
\parbox[t]{15mm}{\multirow{1}{*}{\rotatebox[origin=c]{0}{$\text{GearNet}_{\text{Co-teaching}}$}}}
& $51.95 \pm 1.33$   & $54.37 \pm 1.13$ & $44.49 \pm 1.80$ & $71.87 \pm 0.93$ & $41.79 \pm 1.71$  & $61.45 \pm 0.97$ & $ 54.32 \pm 1.31$\tabularnewline

\parbox[t]{15mm}{\multirow{1}{*}{\rotatebox[origin=c]{0}{$\text{GearNet}_{\text{DANN}}$}}}
&\textbf{59.51 $\pm$ 1.58}   & $61.25 \pm 0.63$ & $41.44 \pm 1.96$ & $72.08 \pm 1.05$ & $44.89 \pm 1.58$  & $69.27 \pm 1.62$ & $58.07 \pm 1.40 $\tabularnewline

\parbox[t]{28mm}{\multirow{1}{*}{\rotatebox[origin=c]{0}{$\text{GearNet}_{\text{TCL}}$}}}
  &$58.85 \pm 0.96$   & \textbf{62.71 $\pm$ 0.73} & \textbf{44.28 $\pm$ 1.53} & \textbf{75.00 $\pm$ 1.67} & \textbf{45.03 $\pm$ 0.85}  & \textbf{75.00 $\pm$ 1.09} & \textbf{60.15 $\pm$ 1.14} \tabularnewline
\bottomrule
% \parbox[t]{15mm}{\multirow{1}{*}{\rotatebox[origin=c]{0}{G+Butterfly}}}
%  &$N$   & $N$ & $N$ & $N$ & $N$  & $N$ & $N$\tabularnewline
% \hline 
% \hline 
\end{tabular}}
\caption{Target accuracy (\%) on Office-31 datasets with Flip-20\% noise. The best results are highlighted in bold.}
\label{flip2}
\par\end{center}
\end{table*}

 \begin{table*}[h]
\small
\begin{center}
\setlength{\tabcolsep}{4pt} % Default value: 6pt
\renewcommand{\arraystretch}{1} % Default value: 1
\label{result}
\resizebox{!}{!}
{ %
\begin{tabular}{l|ccccccc}
\toprule
 Tasks &$A\rightarrow W$  & $A\rightarrow D$  & $W\rightarrow A$  & $W\rightarrow D$  & $D\rightarrow A$  & $D\rightarrow W$  & $\text{Average}$ \tabularnewline
\midrule
\parbox[t]{17mm}{\multirow{1}{*}{\rotatebox[origin=c]{0}{Standard}}}
 & $35.93 \pm 1.09$ & $37.50 \pm 1.14$ & $30.39 \pm 1.21$ & $53.96 \pm 1.04$ & $29.83 \pm 1.81$ & $46.88 \pm 1.35$ & $39.08 \pm 1.27$\tabularnewline
% \midrule

\parbox[t]{17mm}{\multirow{1}{*}{\rotatebox[origin=c]{0}{Co-teaching}}}
  &$35.94 \pm 1.97$  & $37.92 \pm 2.48$ & $28.13 \pm 2.09$ & $49.17 \pm 1.35$ & $26.03 \pm 1.71$ & $37.76 \pm 1.95$ & $35.82 \pm 1.93$\tabularnewline
% \midrule

 \parbox[t]{17mm}{\multirow{1}{*}{\rotatebox[origin=c]{0}{JoCoR}}}
&$36.07 \pm 1.85$   & $37.29 \pm 1.27$ & $27.73 \pm 1.88$ & $48.12 \pm 2.08$ & $26.35 \pm 1.11$  & $38.28 \pm 1.60$ & $35.64 \pm 1.63$ \tabularnewline

% \midrule

\parbox[t]{17mm}{\multirow{1}{*}{\rotatebox[origin=c]{0}{DAN}}}
 & $43.49 \pm 1.17$   & $41.46 \pm 2.14$ & $30.61 \pm 1.43$ & $53.54 \pm 2.01$ & $28.94 \pm 2.36$  & $50.39 \pm 2.35$ & $41.41 \pm 1.91$\tabularnewline

% \midrule
\parbox[t]{17mm}{\multirow{1}{*}{\rotatebox[origin=c]{0}{DANN}}}
 &$36.98 \pm 1.84$   & $40.42 \pm 1.89$ & $30.26 \pm 2.04$ & $53.33 \pm 2.58$ & $30.36 \pm 1.52$  & $44.66 \pm 1.92$ & $39.96 \pm 1.96$\tabularnewline

% \midrule
\parbox[t]{17mm}{\multirow{1}{*}{\rotatebox[origin=c]{0}{TCL}}}
 & $44.79 \pm 0.98$   & $43.75 \pm 1.57$ & $30.82 \pm 1.37$ & $54.17 \pm 1.90$ & $29.97 \pm 1.86$  & $45.96 \pm 1.37$ & $41.58 \pm 1.51$\tabularnewline

% \midrule
% \parbox[t]{17mm}{\multirow{1}{*}{\rotatebox[origin=c]{0}{RDA}}}
% &$N$   & $N$ & $N$ & $N$ & $N$  & $N$ & $N$\tabularnewline
% \hline

% \parbox[t]{17mm}{\multirow{1}{*}{\rotatebox[origin=c]{0}{Butterfly}}}
% &$N$   & $N$ & $N$ & $N$ & $N$  & $N$ & $N$\tabularnewline

\midrule
% \midrule
\parbox[t]{15mm}{\multirow{1}{*}{\rotatebox[origin=c]{0}{$\text{GearNet}_{\text{Co-teaching}}$}}}
& $40.36 \pm 0.85$   & $38.75 \pm 1.14$ & $30.39 \pm 0.58$ & $51.04 \pm 1.17$ & $27.37 \pm 0.64$  & $40.23 \pm 1.36$ & $38.02 \pm 0.96$\tabularnewline

\parbox[t]{17mm}{\multirow{1}{*}{\rotatebox[origin=c]{0}{$\text{GearNet}_{\text{DANN}}$}}}
&$42.45 \pm 1.33$   & $42.08 \pm 1.54$ & \textbf{31.21 $\pm$ 1.73} & \textbf{54.38 $\pm$ 1.48} & \textbf{31.35 $\pm$ 0.79}  & $45.96 \pm 1.63$ & $41.24 \pm 1.42 $\tabularnewline

\parbox[t]{28mm}{\multirow{1}{*}{\rotatebox[origin=c]{0}{$\text{GearNet}_{\text{TCL}}$}}}
&\textbf{48.69 $\pm$ 1.02}   & \textbf{46.25 $\pm$ 1.46} & $30.64 \pm 1.58$ & $53.96 \pm 1.24$ & $31.00 \pm 0.89$  & \textbf{47.79 $\pm$ 1.24} & \textbf{43.06 $\pm$ 1.24} \tabularnewline
\bottomrule
% \parbox[t]{17mm}{\multirow{1}{*}{\rotatebox[origin=c]{0}{G+Butterfly}}}
% &$N$   & $N$ & $N$ & $N$ & $N$  & $N$ & $N$\tabularnewline
% \hline 
% \hline 
\end{tabular}}
\caption{Target accuracy (\%) on Office-31 datasets with Flip-40\% noise. Bold numbers are superior results.}
\label{flip4}
\par\end{center}
\end{table*}

 \begin{table*}[h]
\small
\begin{center}
\renewcommand{\arraystretch}{1} % Default value: 1
\label{result}
\resizebox{\textwidth}{!}{
\setlength{\tabcolsep}{6pt}{ % Default value: 6pt
\begin{tabular}{l|cccccc|ccc}
\toprule
 Tasks &Standard  &Co-teaching  &JoCoR   &DAN  &DANN &TCL  & $\text{GearNet}_{\text{Co-teaching}}$ & $\text{GearNet}_{\text{DANN}}$   & $\text{GearNet}_{\text{TCL}}$ \tabularnewline
\midrule
\parbox[t]{17mm}{\multirow{1}{*}{\rotatebox[origin=c]{0}{$Ar\rightarrow CI$}}}
 &24.51$\pm$1.82 &26.49$\pm$1.40 &26.60$\pm$1.93 &23.12$\pm$0.90 &26.41$\pm$1.51 &26.48$\pm$0.93 &27.82$\pm$0.42 &27.30$\pm$1.07 &\textbf{28.01$\pm$0.75} \tabularnewline
% \midrule

\parbox[t]{17mm}{\multirow{1}{*}{\rotatebox[origin=c]{0}{$Ar\rightarrow Pr$}}}
  &42.41$\pm$1.59  &45.15$\pm$1.79 &45.08$\pm$2.02 &33.94$\pm$0.95 &41.13$\pm$1.20 &42.75$\pm$1.06 &\textbf{52.15$\pm$0.73} &47.01$\pm$1.05 &46.58$\pm$0.93\tabularnewline
% \midrule

 \parbox[t]{17mm}{\multirow{1}{*}{\rotatebox[origin=c]{0}{$Ar\rightarrow Rw$ }}}
&48.75$\pm$1.89 &51.51$\pm$1.67 &51.56$\pm$2.28 &50.33$\pm$1.10 &49.60$\pm$1.66 &49.63$\pm$1.12 &\textbf{54.71$\pm$0.70} &51.12$\pm$1.15 &52.33$\pm$1.50\tabularnewline

% \midrule

\parbox[t]{17mm}{\multirow{1}{*}{\rotatebox[origin=c]{0}{$CI\rightarrow Ar$}}}
 &27.58$\pm$1.60 &27.70$\pm$1.17 &27.91$\pm$1.76 &26.11$\pm$0.65 &34.08$\pm$1.67 &36.22$\pm$0.96  &31.20$\pm$0.77 &36.34$\pm$0.82 &\textbf{37.37$\pm$1.47}\tabularnewline

% \midrule
\parbox[t]{17mm}{\multirow{1}{*}{\rotatebox[origin=c]{0}{$CI\rightarrow Pr$}}}
 &34.60$\pm$1.09 &35.39$\pm$1.63 &34.98$\pm$1.48 &31.61$\pm$1.18 &38.22$\pm$2.21 &41.89$\pm$1.30 &42.77$\pm$0.75  &43.09$\pm$0.98 &\textbf{43.22$\pm$0.91}\tabularnewline

% \midrule
\parbox[t]{17mm}{\multirow{1}{*}{\rotatebox[origin=c]{0}{$CI\rightarrow Rw$}}}
&37.59$\pm$1.45 &26.08$\pm$1.07 &37.20$\pm$1.43 &34.28$\pm$1.40 &42.41$\pm$1.63 &45.35$\pm$1.48  &43.01$\pm$0.84 &44.78$\pm$0.56 &\textbf{46.48$\pm$0.86}\tabularnewline

% \midrule
% \parbox[t]{17mm}{\multirow{1}{*}{\rotatebox[origin=c]{0}{RDA}}}
% &$N$   & $N$ & $N$ & $N$ & $N$  & $N$ & $N$\tabularnewline
% \hline

% \parbox[t]{17mm}{\multirow{1}{*}{\rotatebox[origin=c]{0}{Butterfly}}}
% &$N$   & $N$ & $N$ & $N$ & $N$  & $N$ & $N$\tabularnewline

% \midrule
\parbox[t]{15mm}{\multirow{1}{*}{\rotatebox[origin=c]{0}{$Pr\rightarrow Ar$}}}
&29.33$\pm$0.83 &32.12$\pm$1.31 &32.08$\pm$1.34 &25.12$\pm$0.57 &34.62$\pm$1.71 &35.15$\pm$1.40 &33.92$\pm$1.22  &36.75$\pm$0.83 &\textbf{37.58$\pm$1.27}\tabularnewline
% \midrule
\parbox[t]{15mm}{\multirow{1}{*}{\rotatebox[origin=c]{0}{$Pr\rightarrow CI$ }}}
&23.18$\pm$1.68 &23.92$\pm$0.60 &24.11$\pm$1.59 &23.47$\pm$1.78 &24.00$\pm$1.70 &25.79$\pm$0.69 &23.72$\pm$1.13 &24.11$\pm$1.10 &\textbf{28.46$\pm$1.42}\tabularnewline

\parbox[t]{15mm}{\multirow{1}{*}{\rotatebox[origin=c]{0}{$Pr\rightarrow Rw$ }}}
&46.07$\pm$1.91 &48.32$\pm$1.07 &48.80$\pm$1.39 &40.92$\pm$1.00 &49.65$\pm$1.56 &52.68$\pm$1.55  &50.22$\pm$0.69 &50.89$\pm$1.05 &\textbf{54.84$\pm$1.11}\tabularnewline

\parbox[t]{15mm}{\multirow{1}{*}{\rotatebox[origin=c]{0}{$Rw\rightarrow Ar$}}}
&41.37$\pm$1.58 &43.20$\pm$1.35 &44.03$\pm$1.84 &35.48$\pm$1.09 &43.66$\pm$2.14 &44.98$\pm$0.84  &44.37$\pm$0.91 &44.61$\pm$0.56 &\textbf{46.54$\pm$1.49}\tabularnewline

\parbox[t]{15mm}{\multirow{1}{*}{\rotatebox[origin=c]{0}{$Rw\rightarrow CI$}}}
&26.76$\pm$1.89 &28.12$\pm$1.36 &28.55$\pm$1.59 &27.14$\pm$0.73 &28.30$\pm$1.10 &29.03$\pm$1.92  &28.54$\pm$1.12 &28.80$\pm$1.03 &\textbf{31.06$\pm$1.38}\tabularnewline

\parbox[t]{15mm}{\multirow{1}{*}{\rotatebox[origin=c]{0}{$Rw\rightarrow Pr$}}}
&51.76$\pm$1.71 &53.94$\pm$1.58 &53.87$\pm$1.29 &46.15$\pm$1.46 &51.88$\pm$1.79 &54.84$\pm$1.06  &55.19$\pm$1.27 &53.98$\pm$1.27 &\textbf{57.24$\pm$1.16}\tabularnewline
\midrule
\parbox[t]{15mm}{\multirow{1}{*}{\rotatebox[origin=c]{0}{Average}}}
&36.16$\pm$1.59 &36.83$\pm$1.34 &37.90$\pm$1.65 &33.14$\pm$1.05 &38.66$\pm$1.20 &40.40$\pm$1.19 &40.64$\pm$0.87 &40.73$\pm$0.96 &\textbf{42.48$\pm$1.19}
\tabularnewline
\bottomrule
\end{tabular}}}
\caption{Target accuracy (\%) on Office-Home datasets with Unif-20\% noise. The best results are highlighted in bold.}
\label{home1}
\par\end{center}
\end{table*}

\begin{figure}[t]
\begin{subfigure}{.48\textwidth}
\centering
\subfloat[]{\includegraphics[width=.47\linewidth]{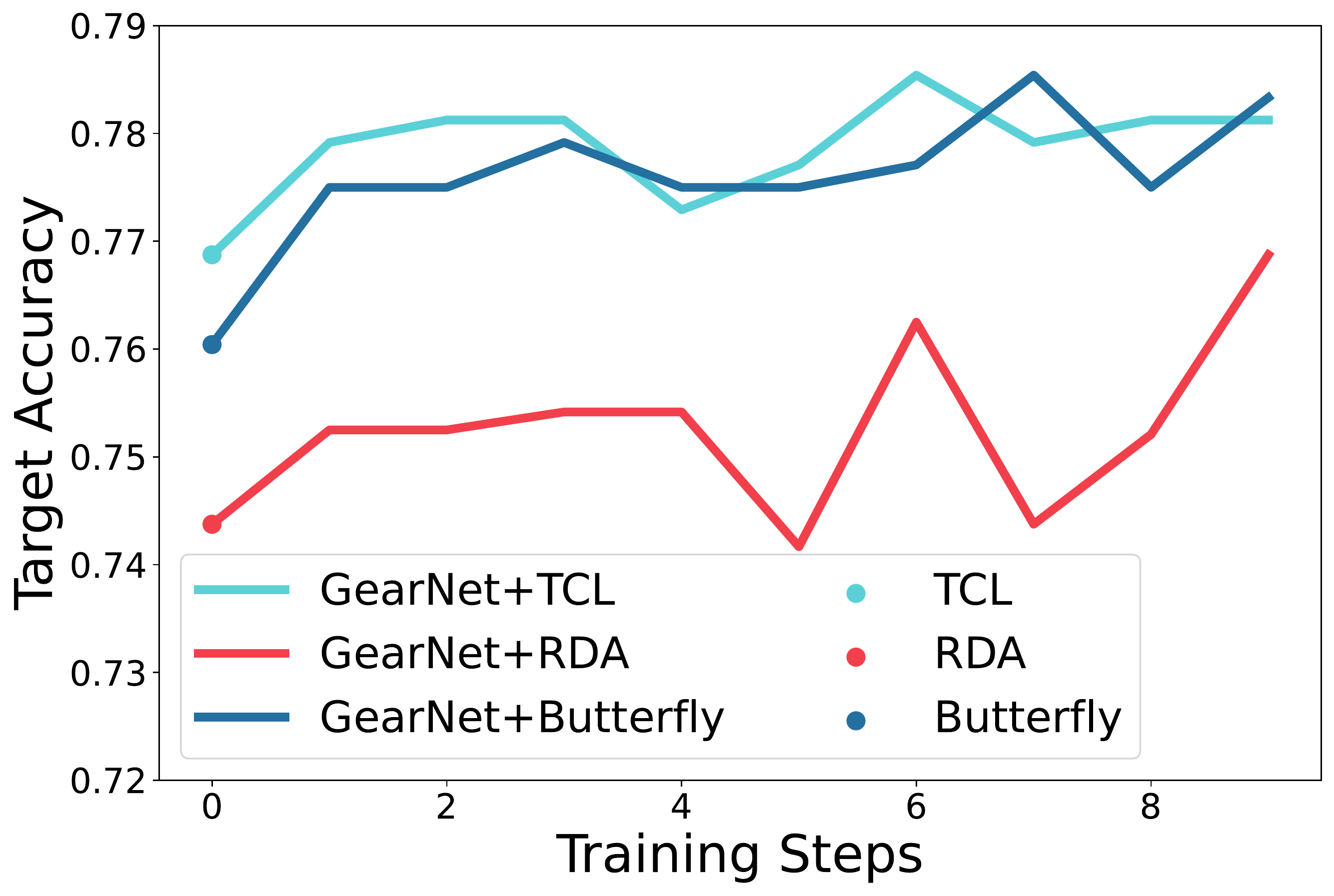}}\quad
\subfloat[]{\includegraphics[width=.47\linewidth]{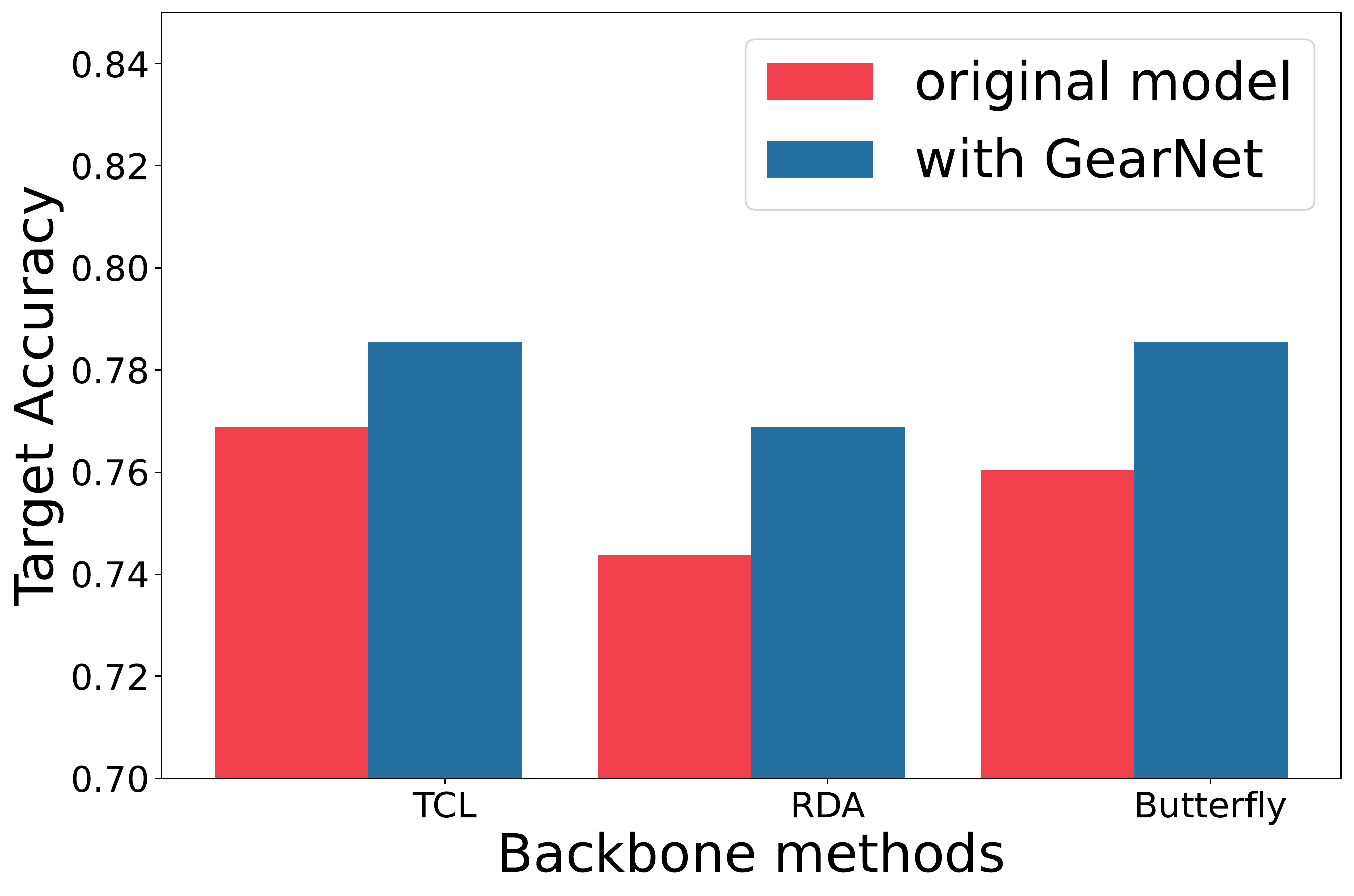}}
\end{subfigure}
\caption{Universal capability of GearNet on $W \rightarrow D$ with Unif-20\% noise for three different WSDA (TCL, RDA, Butterfly) backbone methods. (a) Trend of target accuracy across training steps. (b) Best target accuracy across training steps.}
\label{fig:ablation}
% \vspace{-1.5em}
\end{figure}
\subsection{Experimental setup}
We compare \textbf{GearNet}\footnote{The code is published on \url{https://github.com/Renchunzi-Xie/GearNet.git}} with 6 state-of-the-art baselines, implement all methods by PyTorch, and conduct all the experiments on NVIDIA Tesla V100 GPU. Their details are as follows:
% Standard \cite{he2016deep}, DANN \cite{ganin2016domain}, DAN \cite{long2015learning}, Co-teaching \cite{han2018co}, JoCoR \cite{Wei_2020_CVPR}, and TCL \cite{shu2019transferable}. Among them, there is a simple classification neural network (Standard), two de-noise algorithms (Co-teaching, JoCoR), two domain adaptation algorithms (DAN, DANN) and one weakly-supervised domain adaptation method (TCL). Their details are as follows. 
\textbf{Standard} \cite{he2016deep}, which is a neural network classifier constructed by the pre-trained ResNet-50. Note that ResNet-50 is also used as the feature extractor of the following benchmarks for comparability. \textbf{DANN} \cite{ganin2016domain}, which is constructed by a feature extractor, a classification layer and a domain discriminator. The feature extractor generates a feature space that can confuse the domain discriminator, so that it can decrease the domain discrepancy. \textbf{DAN} \cite{long2015learning}, which proposes MK-MMD as the domain discrepancy measure to reduce the difference between the two domains. \textbf{Co-teaching} \cite{han2018co}, which trains and cross-updates two peer networks simultaneously to combat label noise. \textbf{JoCoR} \cite{Wei_2020_CVPR}, which trains two neural networks simultaneously and calculates a joint loss with Co-regularization to merge their outputs between the two neural networks in order to improve the model robustness against label noise. \textbf{TCL} \cite{shu2019transferable}, which selects transferable and noiseless samples from the source domain to train a model with the same structure as DANN to handle the WSDA issue. \textbf{RDA} \cite{han2020towards}, which proposes an offline curriculum learning to select clean samples from the source domain and a proxy margin discrepancy to eliminate the negative impact of label noise. \textbf{Butterfly} \cite{liu2019butterfly}, which picks clean samples from both of the two domains to train two Co-teaching models simultaneously.

We simulate experiments based on \textbf{Office-31} \cite{saenko2010adapting} and \textbf{Office-Home} \cite{venkateswara2017deep}. The first dataset is a classical dataset for domain adaptation, which contains 4,652 images with 31 classes. Three various domains are contained in the dataset: Amazon (\textbf{A}), Webcam (\textbf{W}) and DSLR (\textbf{D}). They represent that images are collected from amazon.com, web camera and digital SLR camera, respectively. The second dataset consists of 4 domains: Artistic (\textbf{Ar}), Clip Art (\textbf{Cl}), Product (\textbf{Pr}) and Real-World (\textbf{Rw}) containing 15,500 images with 65 classes. They represent different image styles that are artistic depictions, clipart images, images without background and pictures captured by cameras, respectively. We manually inject two types of noisy labels for the source domain: uniform noise \cite{zhang2020distilling} and asymmetry flipping noise \cite{patrini2017making}, and their details are in the supplemental document. 

% We simulate experiments based on \textbf{Office-31} \cite{saenko2010adapting} and \textbf{Office-Home} \cite{venkateswara2017deep}. The first dataset contains 4,652 images with 31 classes and three various domains: Amazon (\textbf{A}), Webcam (\textbf{W}) and DSLR (\textbf{D}). The second dataset consists of 4 domains: Artistic (\textbf{Ar}), Clip Art (\textbf{Cl}), Product (\textbf{Pr}) and Real-World (\textbf{Rw}) containing 15,500 images with 65 classes. We manually inject two types of noisy labels for the source domain: uniform noise \cite{zhang2020distilling} and asymmetry flipping \cite{patrini2017making}.

For comparability, all the experiments use Stochastic gradient descent optimizer with an initial learning rate of 0.003 and a momentum of 0.9. The batch size is set as 32 and the total number of epochs is 200. For GearNet, the total number of steps is set as 10. To measure the performance, we evaluate the target accuracy by all the samples from the target domain, i.e., \emph{target accuracy = {(\# of correct target predictions)/(\# of target domain data)}}. All the experiments are repeated 5 times with different seeds, and we report the average accuracy and their standard deviation on tables.

% All experiments are repeated five times, and we report the average accuracy and their standard deviation. 

\subsection{Numerical results}
\textbf{Results on Office-31.} Table \ref{unif2}, Table \ref{unif4}, Table \ref{flip2} and Table \ref{flip4} report the target-domain accuracy in 6 digit tasks that are combined in pairs by the three domains from Office-31 under different types and levels of noise. 
% The 6 tasks are $Amazon \rightarrow Webcam$ ($A \rightarrow W$), $Amazon \rightarrow DSLR$ ($A \rightarrow D$), $Webcam \rightarrow Amazon$ (W \rightarrow A), $Webcam \rightarrow DSLR$, $DSLR \rightarrow Amazon$ ($D \rightarrow A$) and $DSLR \rightarrow Webcam$ ($D \rightarrow W$). 
% The final results show that GearNet's accuracy is higher than all the baselines. Specifically, the GearNet strategy makes performances of all the WSDA algorithms improve compared with their original results.

Table \ref{unif2} and Table \ref{flip2} represent two simpler cases, since their noise rate is 20\%. The two tables illustrate that our method is able to improve existing backbone methods significantly when the noise rate is low.
% The most significant improvement happens in $D \rightarrow W$ under flip noise from 70.96\% to 75.00\%. The accuracy of every task also increases about 2\%. 
Table \ref{unif4} and Table \ref{flip4} represent two harder cases with 40\% noise rate. The two tables show that GearNet can enhance the performance of original algorithms on most tasks when the noise rate is high. GearNet obtains comparable results on the tasks of $D \rightarrow A$ and $W \rightarrow A$. The reason is that Webcam and DSLR are two small datasets compared with Amazon, so that it is ineffective to transfer knowledge from Webcam or DSLR to Amazon under 40\% noise rate. When we explore pseudo supervision information from Amazon based on the pseudo labels provided by the two small datasets, it is more likely to explore negative information and transfer it back to Amazon. 
% As a result, when we explore Amazon with low-accuracy pseudo labels, it cannot provide useful information for Webcam even though it has the support of the Webcam information from the last model. 
However, when the noise rate becomes 20\%, all the tasks can be improved by GearNet, since all of the source domains are capable to provide adequate transferable information to their target domains and vise versa.

To show the universal capability of GearNet on backbone methods from different fields, we draw Fig. \ref{fig:20}. The left figure illustrates the trend of target accuracy of GearNet across training steps under the three backbone methods. The step 0 refers to the pretrained process, the even number step (i.e., step 2, 4, ...) refers the forward step from the source domain to the target domain, and the odd number step (i.e., step 1, 3, 5, ...) means the backward step from the target domain to the source domain. Specifically, the value of the first point denotes the target accuracy of the original backbone model. The right figure illustrates the performance improvement before and after we incorporate GearNet with the three methods. From the figures, we can observe that GearNet can continuously enhance the performance of the backbone methods. The figures on $D \rightarrow W$ with Unif-40\% noise, Flip-20\% noise and Flip-40\% noise are in the supplemental document.
% \begin{table}[htb]
% \centering
% \setlength{\tabcolsep}{6pt}{ % Default value: 6pt
% \renewcommand{\arraystretch}{1} % Default value: 1
% \resizebox{0.46\textwidth}{!}{ %
% \begin{tabular}{ccccc}
% \toprule 
% &\multicolumn{1}{c}{$\text{GearNet}_{\text{Co-teaching}}$} &\multicolumn{1}{c}{$\text{GearNet}_{\text{DANN}}$} &\multicolumn{1}{c}{$\text{GearNet}_{\text{TCL}}$}\tabularnewline
% \midrule
%   w/o regularization &36.58 &49.09 &49.47  \tabularnewline
% \midrule 
% w/ regularization &\textbf{38.41} &\textbf{54.17} &\textbf{52.47} \tabularnewline
% % \midrule 
% % $\text{GearNet}_{\text{CoTeaching}}$  &$69.14$ & $\textbf{70.57}$ & $\textbf{38.41}$ & $36.58$ \tabularnewline
% % \midrule 
% % $\text{Gearnet}_{\text{DANN}}$  &$\textbf{72.39}$ & $70.18$ & $\textbf{54.17}$ & $49.09$ \tabularnewline
% \bottomrule
% \end{tabular}
% }}
% \caption{Results of ablation study to analyze the importance of the consistency regularization (i.e., $\ell_{guide}$ and $\tilde{\ell}_{guide}$) on $D \rightarrow W$ with Unif-40\% noise. The best results are highlighted in bold.}
% \label{ablation}
% % \vspace{-1.5em}
% \end{table}
\begin{table}[htb]
\centering
\setlength{\tabcolsep}{6pt}{ % Default value: 6pt
\renewcommand{\arraystretch}{1} % Default value: 1
\resizebox{0.46\textwidth}{!}{ %
\begin{tabular}{c|cc|cc|cc}
\toprule 
&\multicolumn{2}{c|}{$\text{GearNet}_{\text{Co-teaching}}$} &\multicolumn{2}{c|}{$\text{GearNet}_{\text{DANN}}$} &\multicolumn{2}{c}{$\text{GearNet}_{\text{TCL}}$}\tabularnewline
\midrule
 Tasks &w/o &w/ &w/o &w/ &w/o &w/  \tabularnewline
 \midrule 
$D\rightarrow W \quad \text{Unif-20}\%$ &67.58 &\textbf{69.14} &70.18 &\textbf{72.39} &71.74 &\textbf{73.44} \tabularnewline
\midrule
 $D\rightarrow W \quad \text{Unif-40}\%$ &36.58 &\textbf{38.41} &49.09 &\textbf{54.17} &49.47 &\textbf{52.47} \tabularnewline
 \midrule 
$Ar\rightarrow Cl \quad \text{Unif-20}\%$ &27.39 &\textbf{27.82} &26.53 &\textbf{27.30} &26.41 &\textbf{28.01} \tabularnewline
\midrule
 $Ar\rightarrow Cl \quad \text{Unif-40}\%$ &18.65 &\textbf{18.75} &17.56 &\textbf{17.58} &16.30 &\textbf{17.93} \tabularnewline
\bottomrule
\end{tabular}
}}
\caption{Results of ablation study 
%to analyze the importance of the consistency regularization (i.e., $\ell_{guide}$ and $\tilde{\ell}_{guide}$) 
on various tasks with Uniform noise. The best results are highlighted in bold.}
\label{ablation}
% \vspace{-1em}
\end{table}
To illustrate the universal capability of GearNet on different WSDA methods, we incorporate three WSDA methods with GearNet including TCL \cite{shu2019transferable}, Butterfly \cite{liu2019butterfly} and RDA \cite{han2020towards} based on the Office-31 benchmark dataset (source: Webcam; target: DSLR). Their results are shown in Fig. \ref{fig:ablation}. The two figures illustrate that GearNet can also significantly improve the prediction accuracy on the target domain for various WSDA methods. 
% The three methods represent the de-noising approach, the domain adaptation approach and the weakly-supervised domain adaptation approach, respectively. 
% From the figures, we can observe that GearNet can continuously enhance the performance of the backbone methods. Fig. \ref{fig:20} (b) illustrates the performance improvement after we incorporate GearNet with the three methods. From this figure, we 

% This figure also shows that GearNet will improve more model classification capability when the model is not qualified to handle weakly-supervised domain adaptation issues, which is also illustrated in Fig \ref{best}. From the second figure, \emph{GearNet+Co-teaching} has a remarkable increase compared with \emph{GearNet+TCL}, while the results of Co-teaching is much lower than TCL. 

% This phenomenon is possible to be caused by the degree of model capability improvement. For Co-teaching method, the increased accuracy includes additional domain adaptation capability and enhanced denoise capability, and the increased accuracy of DANN contains the additional denoise capability and enhanced domain adaptation capability. But for TCL, it only contains the enhanced domain adaptation capability and denoise capability. Meanwhile, the reason why the result of \emph{GearNet+TCL} is still the best followed by the result of \emph{GearNet+DANN} and \emph{GearNet+Co-teaching} is that the model performance is limited by its upper classification level.    

\noindent\textbf{Results on Office-Home.}\quad Table \ref{home1} report the average target accuracy on 12 tasks that are combined in pairs by the 4 domains from Office-Home when the uniform noise rate is 20\%. This table shows that  GearNet can significantly improve the performance compared with original models. Especially, in the case of $Ar\rightarrow Pr$, the target accuracy is from 42\% to 52\% when the backbone method is Co-teaching, which is a significant improvement. 

\section{Ablation study}
% \begin{table}[htb]
% \caption{Results of ablation study to analyze the importance of symmetric KL Divergence on $D \rightarrow W$ with Unif-40\%. Bold numbers are superior results}.
% \label{ablation}
% \centering
% \setlength{\tabcolsep}{6pt}{ % Default value: 6pt
% \renewcommand{\arraystretch}{1} % Default value: 1
% \resizebox{0.46\textwidth}{!}{ %
% \begin{tabular}{c|cc|cc}
% \toprule 
% &\multicolumn{2}{c|}{Unif-20\%} &\multicolumn{2}{c}{Unif-40\%}\tabularnewline
% \midrule
% Method & w/ regularization & w/o regularization & w/ regularization & w/o regularization \tabularnewline
% \midrule 
% $\text{GearNet}_{\text{TCL}}$ & $\textbf{73.44}$ & $71.74$ & $\textbf{52.47}$ & $49.47$  \tabularnewline
% \midrule 
% $\text{GearNet}_{\text{CoTeaching}}$  &$69.14$ & $\textbf{70.57}$ & $\textbf{38.41}$ & $36.58$ \tabularnewline
% \midrule 
% $\text{Gearnet}_{\text{DANN}}$  &$\textbf{72.39}$ & $70.18$ & $\textbf{54.17}$ & $49.09$ \tabularnewline
% \bottomrule
% \end{tabular}
% }}
% % \vspace{-1.5em}
% \end{table}

To illustrate the importance of the consistency regularization (i.e., $\ell_{guide}$ and $\tilde{\ell}_{guide}$), we conduct an ablation study on four tasks, and the results are shown in Table \ref{ablation}. The backbone algorithms are Co-teaching \cite{han2018co}, DANN \cite{ganin2016domain} and TCL \cite{shu2019transferable}, which represent a de-noise method, a domain adaptation method and a WSDA method, respectively. From Table \ref{ablation}, it is clear to see that the guidance on the testing domain from the other model could significantly improve the performance of the classification for most of the methods.

% . Fig. (\ref{fig:ablation}) (a) illustrates that GearNet increases the predictive accuracy of the target domain under different WSDA backbone methods continuously. From Fig. (\ref{fig:ablation}) (b), it illustrates that GearNet can also significantly improve the prediction accuracy on the target domain for various WSDA methods. 

\section{Conclusion}
In this paper, we propose a universal paradigm called GearNet that enhances many existing robust training methods to address the issue of \emph{weakly-supervised domain adaptation}. To the best of our knowledge, our method is the first to explore the benefit of utilizing pseudo supervision knowledge from the target domain in improving the robustness against noisy labels from the source domain. We show that exploring bilateral relationships would further improve the generalization performance when the labels from the source domain are noisy. Extensive experiments show that GearNet is easy to be integrated into existing algorithms and these methods equipped with GearNet can significantly outperform their original performance. Overall, our method is an effective and complementary approach for boosting robustness against noisy labels in the setting of domain adaptation. 

% Different from previous methods, we do not focus on designing delicate models to solve domain adaptation and label noise problems. Instead, we continuously improve the target accuracy by iteratively leveraging the supervision information of both the source domain and the target domain, and enforce the supervision information to guide the performance of the dual model. In the numerical study, we can observe that methods equipped with GearNet can significantly outperform their original performance.    

\section{Acknowledgements}
This research was supported by the National Research Foundation, Singapore under its AI Singapore Programme (AISG Award No: AISG-RP-2019-0013), National Satellite of Excellence in Trustworthy Software Systems (Award No: NSOE-TSS2019-01), and NTU. Lei Feng was supported by the National Natural Science Foundation of China under Grant 62106028 and CAAI-Huawei MindSpore Open Fund.

% \clearpage
% \bibliographystyle{aaai_name}
\bibliography{main}

\begin{thebibliography}{51}
\providecommand{\natexlab}[1]{#1}

\bibitem[{Blitzer, McDonald, and Pereira(2006)}]{blitzer2006domain}
Blitzer, J.; McDonald, R.; and Pereira, F. 2006.
\newblock Domain adaptation with structural correspondence learning.
\newblock In \emph{Proceedings of the 2006 Conference on Empirical Methods in
  Natural Language Processing}, 120--128.

\bibitem[{Chen et~al.(2020)Chen, Wei, Kumar, and Ma}]{chen2020self}
Chen, Y.; Wei, C.; Kumar, A.; and Ma, T. 2020.
\newblock Self-training avoids using spurious features under domain shift.
\newblock \emph{arXiv preprint arXiv:2006.10032}.

\bibitem[{Combes et~al.(2020)Combes, Zhao, Wang, and Gordon}]{combes2020domain}
Combes, R. T.~d.; Zhao, H.; Wang, Y.-X.; and Gordon, G. 2020.
\newblock Domain adaptation with conditional distribution matching and
  generalized label shift.
\newblock \emph{arXiv preprint arXiv:2003.04475}.

\bibitem[{Dong et~al.(2021{\natexlab{a}})Dong, Cong, Sun, Fang, and
  Ding}]{9616392}
Dong, J.; Cong, Y.; Sun, G.; Fang, Z.; and Ding, Z. 2021{\natexlab{a}}.
\newblock Where and How to Transfer: Knowledge Aggregation-Induced
  Transferability Perception for. Unsupervised Domain Adaptation.
\newblock \emph{IEEE Transactions on Pattern Analysis and Machine
  Intelligence}.

\bibitem[{Dong et~al.(2020)Dong, Cong, Sun, Zhong, and
  Xu}]{What_Transferred_Dong_CVPR2020}
Dong, J.; Cong, Y.; Sun, G.; Zhong, B.; and Xu, X. 2020.
\newblock What Can Be Transferred: Unsupervised Domain Adaptation for
  Endoscopic Lesions Segmentation.
\newblock In \emph{IEEE/CVF Conference on Computer Vision and Pattern
  Recognition (CVPR)}, 4022--4031.

\bibitem[{Dong et~al.(2021{\natexlab{b}})Dong, Fang, Liu, Sun, and
  Liu}]{DBLP:conf/NeuriPS/FangLLL021}
Dong, J.; Fang, Z.; Liu, A.; Sun, G.; and Liu, T. 2021{\natexlab{b}}.
\newblock Confident-anchor-induced multi-source-free domain adaptation.
\newblock In \emph{NeurIPS}.

\bibitem[{Fang et~al.(2021{\natexlab{a}})Fang, Lu, Liu, Liu, and
  Zhang}]{DBLP:conf/icml/FangLLL021}
Fang, Z.; Lu, J.; Liu, A.; Liu, F.; and Zhang, G. 2021{\natexlab{a}}.
\newblock Learning Bounds for Open-Set Learning.
\newblock In \emph{ICML}, 3122--3132.

\bibitem[{Fang et~al.(2021{\natexlab{b}})Fang, Lu, Liu, Xuan, and
  Zhang}]{DBLP:journals/tnn/FangLLXZ21}
Fang, Z.; Lu, J.; Liu, F.; Xuan, J.; and Zhang, G. 2021{\natexlab{b}}.
\newblock Open Set Domain Adaptation: Theoretical Bound and Algorithm.
\newblock \emph{{IEEE} Trans. Neural Networks Learn. Syst.}, 4309--4322.

\bibitem[{Feng et~al.(2020)Feng, Shu, Lin, Lv, Li, and An}]{feng2020can}
Feng, L.; Shu, S.; Lin, Z.; Lv, F.; Li, L.; and An, B. 2020.
\newblock Can cross entropy loss be robust to label noise?
\newblock In \emph{International Joint Conference on Artificial Intelligence},
  2206--2212.

\bibitem[{Fr{\'e}nay and Verleysen(2013)}]{frenay2013classification}
Fr{\'e}nay, B.; and Verleysen, M. 2013.
\newblock Classification in the presence of label noise: a survey.
\newblock \emph{IEEE Transactions on Neural Networks and Learning Systems},
  25(5): 845--869.

\bibitem[{Ganin et~al.(2016)Ganin, Ustinova, Ajakan, Germain, Larochelle,
  Laviolette, Marchand, and Lempitsky}]{ganin2016domain}
Ganin, Y.; Ustinova, E.; Ajakan, H.; Germain, P.; Larochelle, H.; Laviolette,
  F.; Marchand, M.; and Lempitsky, V. 2016.
\newblock Domain-adversarial training of neural networks.
\newblock \emph{The Journal of Machine Learning Research}, 17(1): 2096--2030.

\bibitem[{Ghafoorian et~al.(2017)Ghafoorian, Mehrtash, Kapur, Karssemeijer,
  Marchiori, Pesteie, Guttmann, de~Leeuw, Tempany, Van~Ginneken
  et~al.}]{ghafoorian2017transfer}
Ghafoorian, M.; Mehrtash, A.; Kapur, T.; Karssemeijer, N.; Marchiori, E.;
  Pesteie, M.; Guttmann, C.~R.; de~Leeuw, F.-E.; Tempany, C.~M.; Van~Ginneken,
  B.; et~al. 2017.
\newblock Transfer learning for domain adaptation in mri: Application in brain
  lesion segmentation.
\newblock In \emph{International Conference on Medical Image Computing and
  Computer-assisted Intervention}, 516--524. Springer.

\bibitem[{Ghosh, Kumar, and Sastry(2017)}]{ghosh2017robust}
Ghosh, A.; Kumar, H.; and Sastry, P. 2017.
\newblock Robust loss functions under label noise for deep neural networks.
\newblock In \emph{Proceedings of the AAAI Conference on Artificial
  Intelligence}, volume~31, 1919–--1925.

\bibitem[{Goldberger and Ben-Reuven(2016)}]{goldberger2016training}
Goldberger, J.; and Ben-Reuven, E. 2016.
\newblock Training deep neural-networks using a noise adaptation layer.
\newblock In \emph{Proceedings of the 5th International Conference on Learning
  Representation}.

\bibitem[{Han et~al.(2018)Han, Yao, Yu, Niu, Xu, Hu, Tsang, and
  Sugiyama}]{han2018co}
Han, B.; Yao, Q.; Yu, X.; Niu, G.; Xu, M.; Hu, W.; Tsang, I.; and Sugiyama, M.
  2018.
\newblock Co-teaching: Robust training of deep neural networks with extremely
  noisy labels.
\newblock \emph{arXiv preprint arXiv:1804.06872}.

\bibitem[{Han, Luo, and Wang(2019)}]{han2019deep}
Han, J.; Luo, P.; and Wang, X. 2019.
\newblock Deep self-learning from noisy labels.
\newblock In \emph{Proceedings of the IEEE/CVF International Conference on
  Computer Vision}, 5138--5147.

\bibitem[{Han et~al.(2020)Han, Gui, Cui, and Yin}]{han2020towards}
Han, Z.; Gui, X.-J.; Cui, C.; and Yin, Y. 2020.
\newblock Towards Accurate and Robust Domain Adaptation under Noisy
  Environments.
\newblock \emph{arXiv preprint arXiv:2004.12529}.

\bibitem[{He et~al.(2016)He, Zhang, Ren, and Sun}]{he2016deep}
He, K.; Zhang, X.; Ren, S.; and Sun, J. 2016.
\newblock Deep residual learning for image recognition.
\newblock In \emph{Proceedings of the IEEE Conference on Computer Vision and
  Pattern Recognition}, 770--778.

\bibitem[{Hoffman et~al.(2014)Hoffman, Guadarrama, Tzeng, Hu, Donahue,
  Girshick, Darrell, and Saenko}]{hoffman2014lsda}
Hoffman, J.; Guadarrama, S.; Tzeng, E.; Hu, R.; Donahue, J.; Girshick, R.;
  Darrell, T.; and Saenko, K. 2014.
\newblock LSDA: Large scale detection through adaptation.
\newblock \emph{arXiv preprint arXiv:1407.5035}.

\bibitem[{Hoffman et~al.(2018)Hoffman, Tzeng, Park, Zhu, Isola, Saenko, Efros,
  and Darrell}]{hoffman2018cycada}
Hoffman, J.; Tzeng, E.; Park, T.; Zhu, J.-Y.; Isola, P.; Saenko, K.; Efros, A.;
  and Darrell, T. 2018.
\newblock Cycada: Cycle-consistent adversarial domain adaptation.
\newblock In \emph{International Conference on Machine Learning}, 1989--1998.
  PMLR.

\bibitem[{Jiang et~al.(2018)Jiang, Zhou, Leung, Li, and
  Fei-Fei}]{jiang2018mentornet}
Jiang, L.; Zhou, Z.; Leung, T.; Li, L.-J.; and Fei-Fei, L. 2018.
\newblock Mentornet: Learning data-driven curriculum for very deep neural
  networks on corrupted labels.
\newblock In \emph{International Conference on Machine Learning}, 2304--2313.
  PMLR.

\bibitem[{Kamnitsas et~al.(2017)Kamnitsas, Baumgartner, Ledig, Newcombe,
  Simpson, Kane, Menon, Nori, Criminisi, Rueckert
  et~al.}]{kamnitsas2017unsupervised}
Kamnitsas, K.; Baumgartner, C.; Ledig, C.; Newcombe, V.; Simpson, J.; Kane, A.;
  Menon, D.; Nori, A.; Criminisi, A.; Rueckert, D.; et~al. 2017.
\newblock Unsupervised domain adaptation in brain lesion segmentation with
  adversarial networks.
\newblock In \emph{International Conference on Information Processing in
  Medical Imaging}, 597--609. Springer.

\bibitem[{Kullback and Leibler(1951)}]{kullback1951information}
Kullback, S.; and Leibler, R.~A. 1951.
\newblock On information and sufficiency.
\newblock \emph{The Annals of Mathematical Statistics}, 22(1): 79--86.

\bibitem[{Lee and Raginsky(2017)}]{lee2017minimax}
Lee, J.; and Raginsky, M. 2017.
\newblock Minimax statistical learning with wasserstein distances.
\newblock \emph{arXiv preprint arXiv:1705.07815}.

\bibitem[{Liu et~al.(2019)Liu, Lu, Han, Niu, Zhang, and
  Sugiyama}]{liu2019butterfly}
Liu, F.; Lu, J.; Han, B.; Niu, G.; Zhang, G.; and Sugiyama, M. 2019.
\newblock Butterfly: A panacea for all difficulties in wildly unsupervised
  domain adaptation.
\newblock \emph{arXiv preprint arXiv:1905.07720}.

\bibitem[{Long et~al.(2015)Long, Cao, Wang, and Jordan}]{long2015learning}
Long, M.; Cao, Y.; Wang, J.; and Jordan, M. 2015.
\newblock Learning transferable features with deep adaptation networks.
\newblock In \emph{International Conference on Machine Learning}, 97--105.
  PMLR.

\bibitem[{Luo et~al.(2019)Luo, Li, Zhou, Yang, Chang, Sui, and
  Sun}]{luo2019dual}
Luo, F.; Li, P.; Zhou, J.; Yang, P.; Chang, B.; Sui, Z.; and Sun, X. 2019.
\newblock A dual reinforcement learning framework for unsupervised text style
  transfer.
\newblock \emph{arXiv preprint arXiv:1905.10060}.

\bibitem[{Menon et~al.(2015)Menon, Van~Rooyen, Ong, and
  Williamson}]{menon2015learning}
Menon, A.; Van~Rooyen, B.; Ong, C.~S.; and Williamson, B. 2015.
\newblock Learning from corrupted binary labels via class-probability
  estimation.
\newblock In \emph{International Conference on Machine Learning}, 125--134.
  PMLR.

\bibitem[{Pan et~al.(2010)Pan, Tsang, Kwok, and Yang}]{pan2010domain}
Pan, S.~J.; Tsang, I.~W.; Kwok, J.~T.; and Yang, Q. 2010.
\newblock Domain adaptation via transfer component analysis.
\newblock \emph{IEEE Transactions on Neural Networks}, 22(2): 199--210.

\bibitem[{Pan and Yang(2009)}]{pan2009survey}
Pan, S.~J.; and Yang, Q. 2009.
\newblock A survey on transfer learning.
\newblock \emph{IEEE Transactions on Knowledge and Data Engineering}, 22(10):
  1345--1359.

\bibitem[{Patrini et~al.(2017)Patrini, Rozza, Krishna~Menon, Nock, and
  Qu}]{patrini2017making}
Patrini, G.; Rozza, A.; Krishna~Menon, A.; Nock, R.; and Qu, L. 2017.
\newblock Making deep neural networks robust to label noise: A loss correction
  approach.
\newblock In \emph{Proceedings of the IEEE Conference on Computer Vision and
  Pattern Recognition}, 1944--1952.

\bibitem[{Ren et~al.(2018)Ren, Zeng, Yang, and Urtasun}]{ren2018learning}
Ren, M.; Zeng, W.; Yang, B.; and Urtasun, R. 2018.
\newblock Learning to reweight examples for robust deep learning.
\newblock In \emph{International Conference on Machine Learning}, 4334--4343.
  PMLR.

\bibitem[{Saenko et~al.(2010)Saenko, Kulis, Fritz, and
  Darrell}]{saenko2010adapting}
Saenko, K.; Kulis, B.; Fritz, M.; and Darrell, T. 2010.
\newblock Adapting visual category models to new domains.
\newblock In \emph{European Conference on Computer Vision}, 213--226. Springer.

\bibitem[{Shao, Zhu, and Li(2014)}]{shao2014transfer}
Shao, L.; Zhu, F.; and Li, X. 2014.
\newblock Transfer learning for visual categorization: A survey.
\newblock \emph{IEEE Transactions on Neural Networks and Learning Systems},
  26(5): 1019--1034.

\bibitem[{Shu et~al.(2019)Shu, Cao, Long, and Wang}]{shu2019transferable}
Shu, Y.; Cao, Z.; Long, M.; and Wang, J. 2019.
\newblock Transferable curriculum for weakly-supervised domain adaptation.
\newblock In \emph{Proceedings of the AAAI Conference on Artificial
  Intelligence}, volume~33, 4951--4958.

\bibitem[{Tzeng et~al.(2017)Tzeng, Hoffman, Saenko, and
  Darrell}]{tzeng2017adversarial}
Tzeng, E.; Hoffman, J.; Saenko, K.; and Darrell, T. 2017.
\newblock Adversarial discriminative domain adaptation.
\newblock In \emph{Proceedings of the IEEE Conference on Computer Vision and
  Pattern Recognition}, 7167--7176.

\bibitem[{Venkateswara et~al.(2017)Venkateswara, Eusebio, Chakraborty, and
  Panchanathan}]{venkateswara2017deep}
Venkateswara, H.; Eusebio, J.; Chakraborty, S.; and Panchanathan, S. 2017.
\newblock Deep hashing network for unsupervised domain adaptation.
\newblock In \emph{Proceedings of the IEEE Conference on Computer Vision and
  Pattern Recognition}, 5018--5027.

\bibitem[{Wang and Zheng(2015)}]{wang2015transfer}
Wang, D.; and Zheng, T.~F. 2015.
\newblock Transfer learning for speech and language processing.
\newblock In \emph{2015 Asia-Pacific Signal and Information Processing
  Association Annual Summit and Conference}, 1225--1237. IEEE.

\bibitem[{Wei et~al.(2020)Wei, Feng, Chen, and An}]{Wei_2020_CVPR}
Wei, H.; Feng, L.; Chen, X.; and An, B. 2020.
\newblock Combating Noisy Labels by Agreement: A Joint Training Method with
  Co-Regularization.
\newblock In \emph{Proceedings of the IEEE/CVF Conference on Computer Vision
  and Pattern Recognition}.

\bibitem[{Wu, Zhang, and R{\'e}(2020)}]{wu2020understanding}
Wu, S.; Zhang, H.~R.; and R{\'e}, C. 2020.
\newblock Understanding and improving information transfer in multi-task
  learning.
\newblock \emph{arXiv preprint arXiv:2005.00944}.

\bibitem[{Yang et~al.(2020)Yang, Xia, Ding, and Ding}]{yang2020bi}
Yang, G.; Xia, H.; Ding, M.; and Ding, Z. 2020.
\newblock Bi-directional generation for unsupervised domain adaptation.
\newblock In \emph{Proceedings of the AAAI Conference on Artificial
  Intelligence}, volume~34, 6615--6622.

\bibitem[{Yu et~al.(2020)Yu, Liu, Gong, Zhang, Batmanghelich, and
  Tao}]{yu2020label}
Yu, X.; Liu, T.; Gong, M.; Zhang, K.; Batmanghelich, K.; and Tao, D. 2020.
\newblock Label-noise robust domain adaptation.
\newblock In \emph{International Conference on Machine Learning}, 10913--10924.
  PMLR.

\bibitem[{Zellinger et~al.(2017)Zellinger, Grubinger, Lughofer,
  Natschl{\"a}ger, and Saminger-Platz}]{zellinger2017central}
Zellinger, W.; Grubinger, T.; Lughofer, E.; Natschl{\"a}ger, T.; and
  Saminger-Platz, S. 2017.
\newblock Central moment discrepancy (cmd) for domain-invariant representation
  learning.
\newblock \emph{arXiv preprint arXiv:1702.08811}.

\bibitem[{Zhang et~al.(2016)Zhang, Bengio, Hardt, Recht, and
  Vinyals}]{zhang2016understanding}
Zhang, C.; Bengio, S.; Hardt, M.; Recht, B.; and Vinyals, O. 2016.
\newblock Understanding deep learning requires rethinking generalization.
\newblock \emph{arXiv preprint arXiv:1611.03530}.

\bibitem[{Zhang et~al.(2013)Zhang, Sch{\"o}lkopf, Muandet, and
  Wang}]{zhang2013domain}
Zhang, K.; Sch{\"o}lkopf, B.; Muandet, K.; and Wang, Z. 2013.
\newblock Domain adaptation under target and conditional shift.
\newblock In \emph{International Conference on Machine Learning}, 819--827.
  PMLR.

\bibitem[{Zhang et~al.(2018)Zhang, Xiang, Hospedales, and Lu}]{zhang2018deep}
Zhang, Y.; Xiang, T.; Hospedales, T.~M.; and Lu, H. 2018.
\newblock Deep mutual learning.
\newblock In \emph{Proceedings of the IEEE Conference on Computer Vision and
  Pattern Recognition}, 4320--4328.

\bibitem[{Zhang and Sabuncu(2018)}]{zhang2018generalized}
Zhang, Z.; and Sabuncu, M.~R. 2018.
\newblock Generalized cross entropy loss for training deep neural networks with
  noisy labels.
\newblock In \emph{32nd Conference on Neural Information Processing Systems},
  8778–8788.

\bibitem[{Zhang et~al.(2020)Zhang, Zhang, Arik, Lee, and
  Pfister}]{zhang2020distilling}
Zhang, Z.; Zhang, H.; Arik, S.~O.; Lee, H.; and Pfister, T. 2020.
\newblock Distilling effective supervision from severe label noise.
\newblock In \emph{Proceedings of the IEEE/CVF Conference on Computer Vision
  and Pattern Recognition}, 9294--9303.

\bibitem[{Zhu et~al.(2017)Zhu, Park, Isola, and Efros}]{zhu2017unpaired}
Zhu, J.-Y.; Park, T.; Isola, P.; and Efros, A.~A. 2017.
\newblock Unpaired image-to-image translation using cycle-consistent
  adversarial networks.
\newblock In \emph{Proceedings of the IEEE International Conference on Computer
  Vision}, 2223--2232.

\bibitem[{Zhuang et~al.(2015)Zhuang, Cheng, Luo, Pan, and
  He}]{zhuang2015supervised}
Zhuang, F.; Cheng, X.; Luo, P.; Pan, S.~J.; and He, Q. 2015.
\newblock Supervised representation learning: Transfer learning with deep
  autoencoders.
\newblock In \emph{Twenty-Fourth International Joint Conference on Artificial
  Intelligence}, 4119--4125.

\bibitem[{Zou et~al.(2019)Zou, Yu, Liu, Kumar, and Wang}]{zou2019confidence}
Zou, Y.; Yu, Z.; Liu, X.; Kumar, B.; and Wang, J. 2019.
\newblock Confidence regularized self-training.
\newblock In \emph{Proceedings of the IEEE/CVF International Conference on
  Computer Vision}, 5982--5991.

\end{thebibliography}
\end{document}

\maketitle
\section{A. Figures with Unif-40\%, Flip-20\% and Flip-40\% noise}
\begin{figure}[h]
\begin{subfigure}{.48\textwidth}
\centering
\subfloat[]{\includegraphics[width=.47\linewidth]{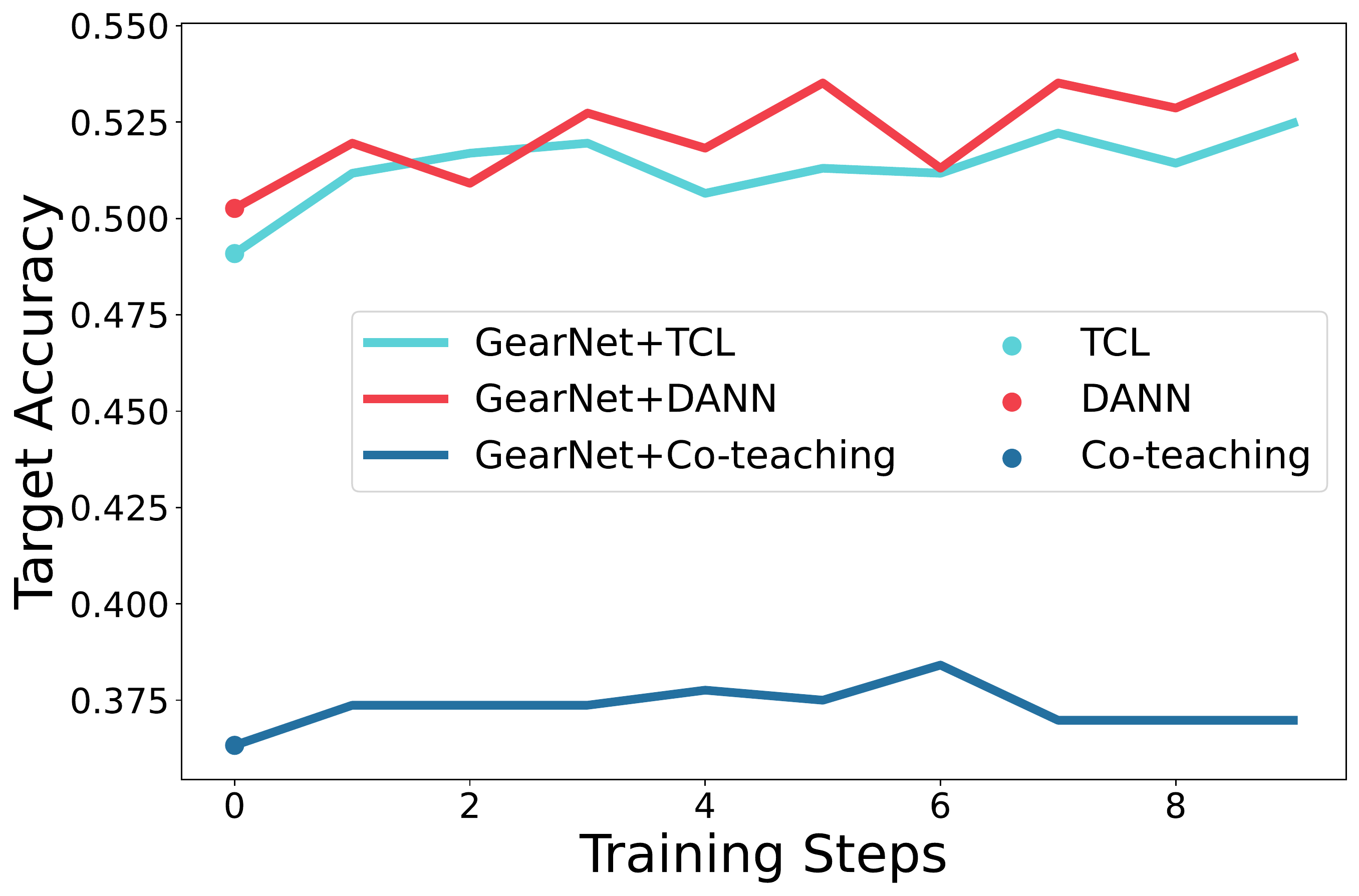}}\quad
\subfloat[]{\includegraphics[width=.47\linewidth]{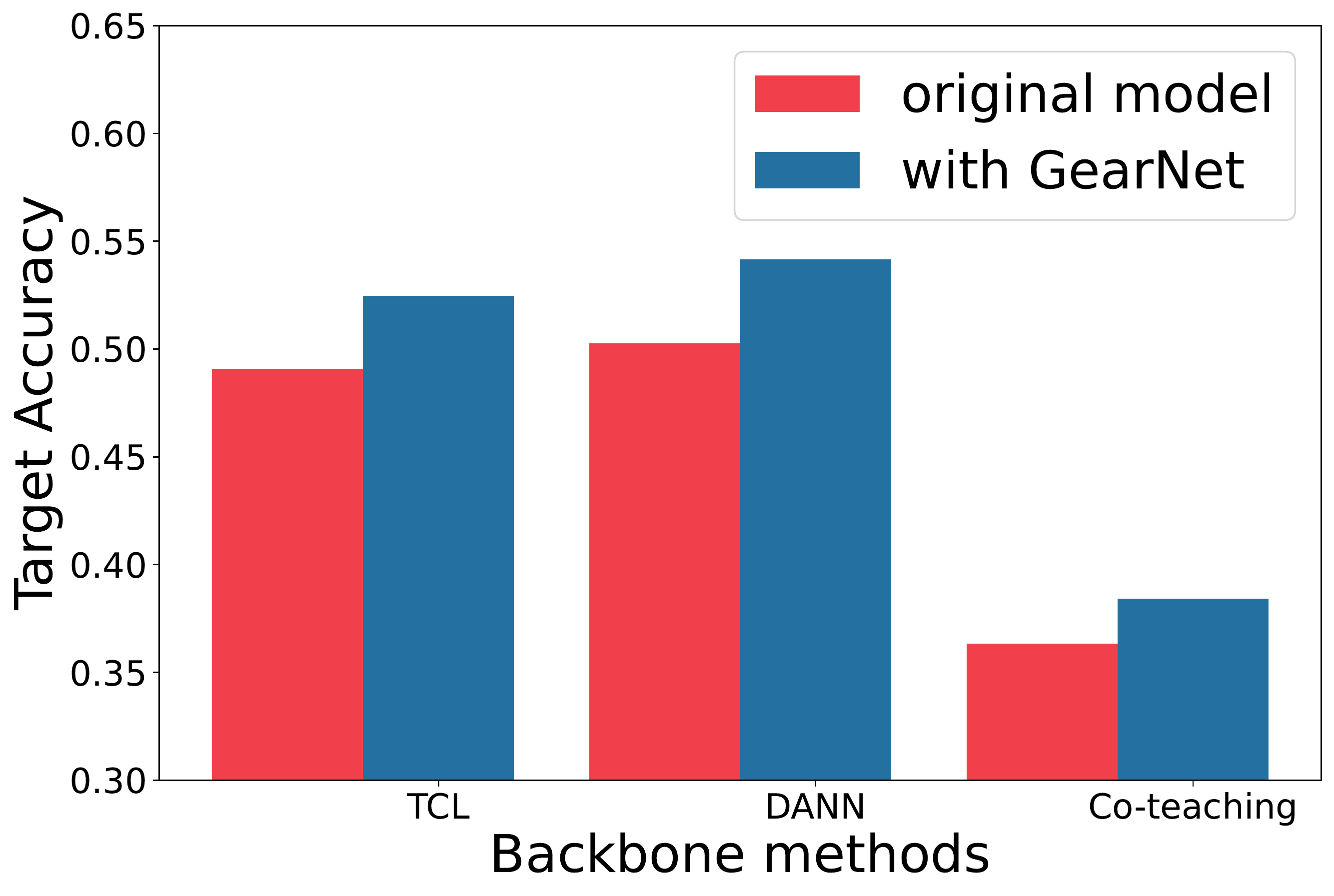}}
\end{subfigure}
\caption{Universal capability of GearNet on $D \rightarrow W$ with Unif-40\% noise under WSDA (TCL), de-noise (Co-teaching) and UDA (DANN) backbone methods. (a) Trend of target accuracy across training steps. (b) Best target accuracy across training steps}
\label{fig:unif-40}
\end{figure}

Fig. \ref{fig:unif-40} illustrates the universal capability of GearNet on $D\rightarrow W$ with Unif-40\% noise when the backbone methods are TCL, Co-teaching and DANN. From the Fig. \ref{fig:unif-40}, we can observe that GearNet improves the performance of backbone methods continuously across training steps.

\begin{figure}[h]
\begin{subfigure}{.48\textwidth}
\centering
\subfloat[]{\includegraphics[width=.47\linewidth]{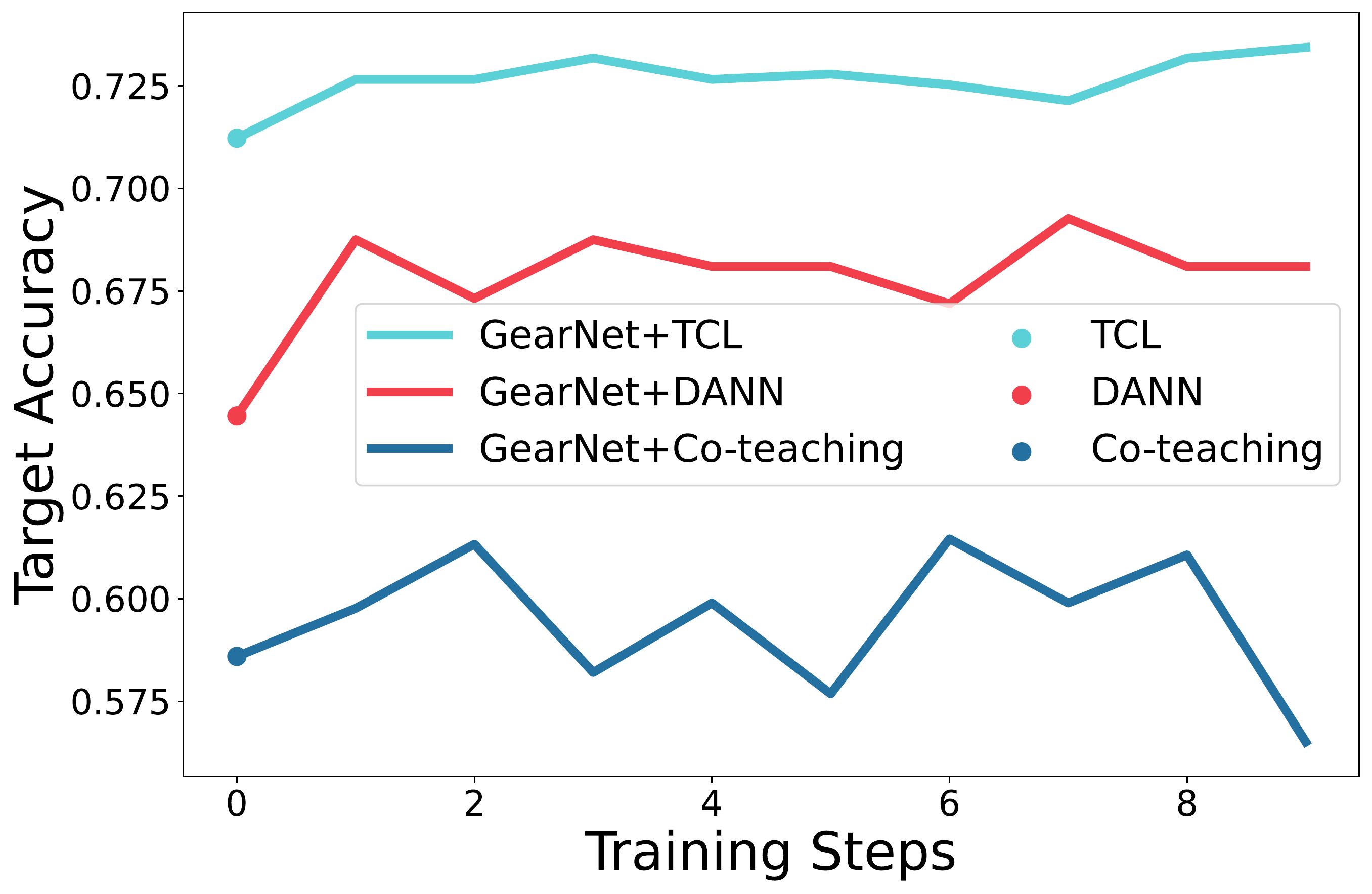}}\quad
\subfloat[]{\includegraphics[width=.47\linewidth]{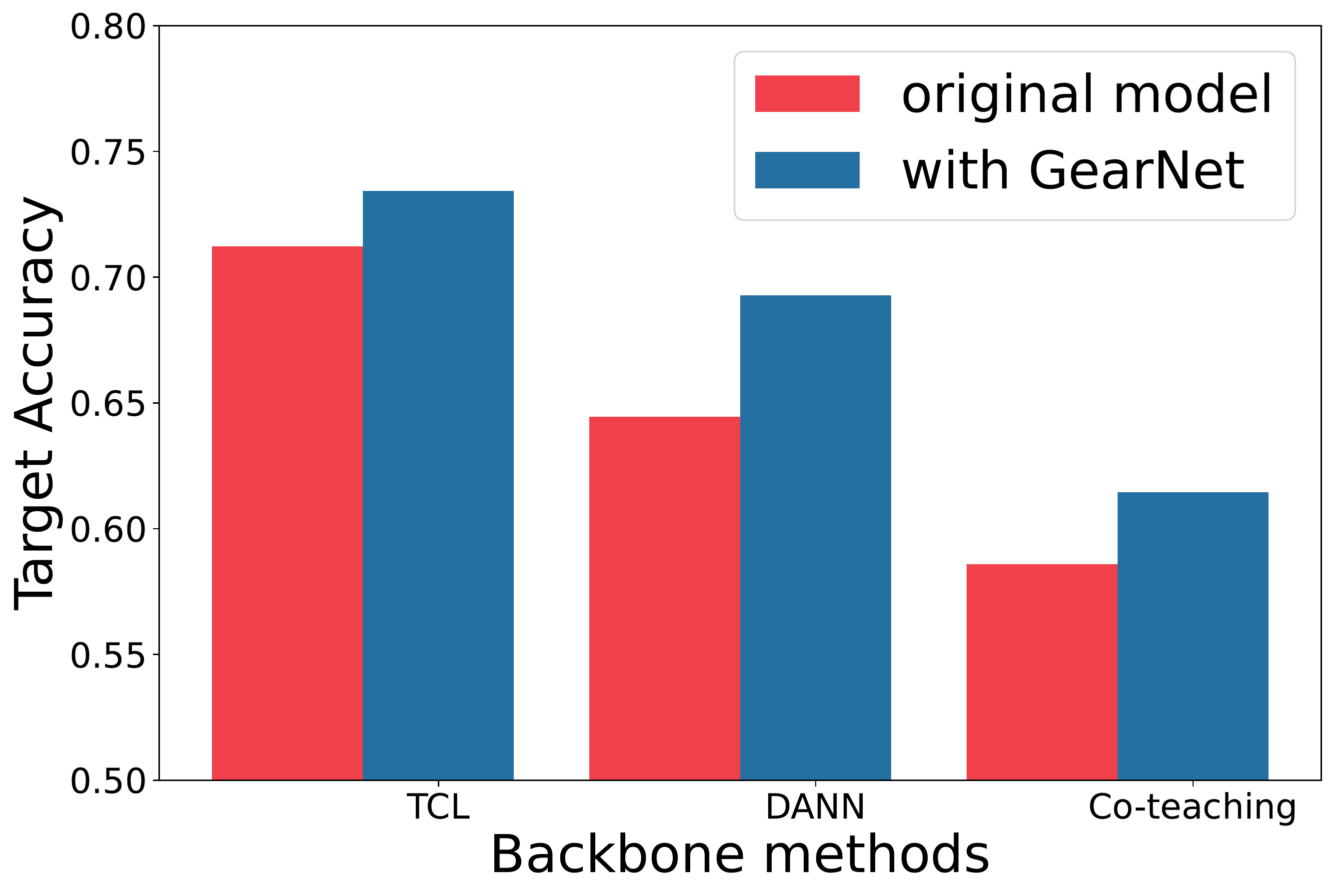}}
\end{subfigure}
\caption{Universal capability of GearNet on $D \rightarrow W$ with Flip-20\% noise under WSDA (TCL), de-noise (Co-teaching) and UDA (DANN) backbone methods. (a) Trend of target accuracy across training steps. (b) Best target accuracy across training steps}
\label{fig:flip-20}
\end{figure}

Fig. \ref{fig:flip-20} illustrates the universal capability of GearNet on $D\rightarrow W$ with Flip-20\% noise when the backbone methods are TCL, Co-teaching and DANN. From the Fig. \ref{fig:flip-20}, we can observe that GearNet improves the best performance of backbone methods. For TCL and DANN, GearNet could continuously improve their performance across training steps. For Co-teaching, GearNet could improve its best performance. 

\begin{figure}[t]
\begin{subfigure}{.48\textwidth}
\centering
\subfloat[]{\includegraphics[width=.47\linewidth]{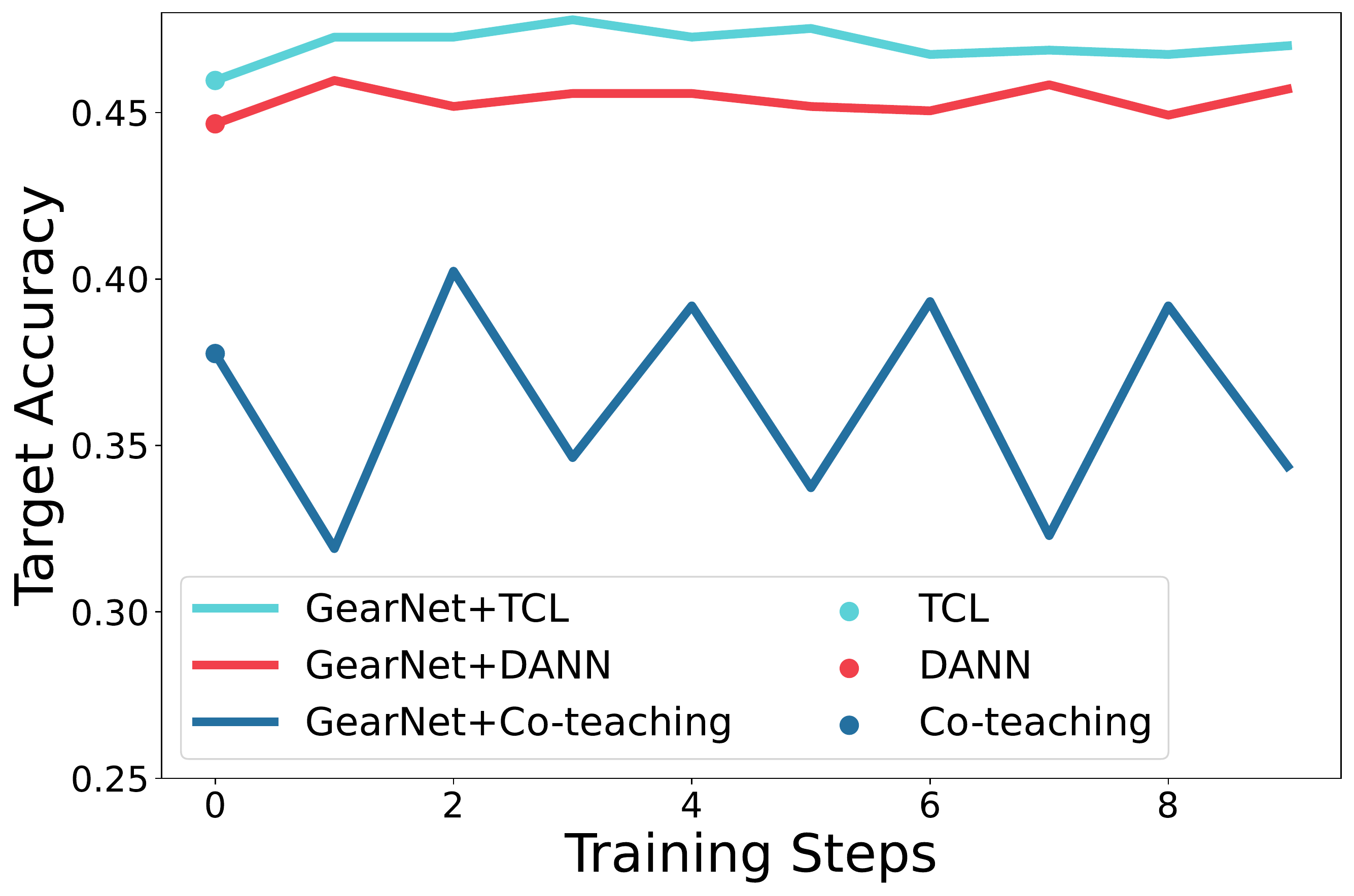}}\quad
\subfloat[]{\includegraphics[width=.47\linewidth]{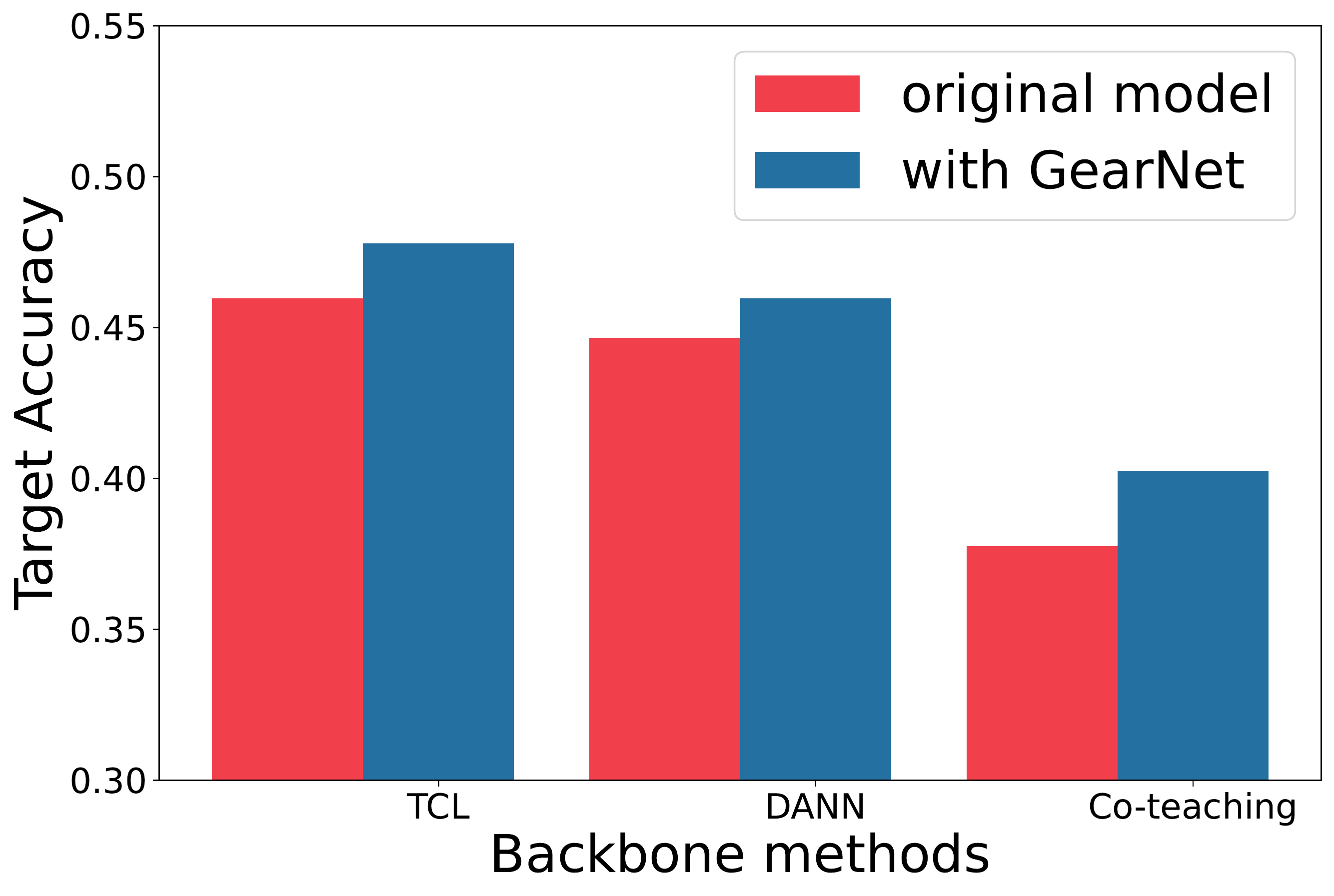}}
\end{subfigure}
\caption{Universal capability of GearNet on $D \rightarrow W$ with Flip-40\% noise under WSDA (TCL), de-noise (Co-teaching) and UDA (DANN) backbone methods. (a) Trend of target accuracy across training steps. (b) Best target accuracy across training steps}
\label{fig:flip-40}
\end{figure}

Fig. \ref{fig:flip-40} illustrates the universal capability of GearNet on $D\rightarrow W$ with Flip-40\% noise when the backbone methods are TCL, Co-teaching and DANN. From the Fig. \ref{fig:flip-40}, we can observe that GearNet improves the best performance of backbone methods. For TCL and DANN, GearNet could continuously improve their performance across training steps. For Co-teaching, GearNet could improve its best performance and its lower performance. 

\section{B. Numerical results on Office-Home with Unif-40\% noise}
Table \ref{home2} reports the average target accuracy on 12 tasks that are combined in pairs by the 4 domains from Office-Home when the uniform noise rate is 40\%. This table shows that  GearNet can significantly improve the performance compared with original models. On average, the performance improvement is 1\% when the noise rate is 40\%.
 \begin{table*}[!t]
\small
\begin{center}
\renewcommand{\arraystretch}{1} % Default value: 1
\label{result}
\resizebox{\textwidth}{!}{
\setlength{\tabcolsep}{6pt}{ % Default value: 6pt
\begin{tabular}{l|cccccc|ccc}
\toprule
 Tasks &Standard  &Co-teaching  &JoCoR   &DAN  &DANN &TCL  & $\text{GearNet}_{\text{Co-teaching}}$ & $\text{GearNet}_{\text{DANN}}$   & $\text{GearNet}_{\text{TCL}}$ \tabularnewline
\midrule
\parbox[t]{17mm}{\multirow{1}{*}{\rotatebox[origin=c]{0}{$Ar\rightarrow CI$}}}
 &15.99$\pm$2.04 &17.53$\pm$1.74 &17.30$\pm$1.30 &14.80$\pm$1.92 &17.11$\pm$0.62 &17.13$\pm$1.11 &\textbf{}{18.75$\pm$1.81} &17.58$\pm$0.95 &17.93$\pm$0.64 \tabularnewline
% \midrule

\parbox[t]{17mm}{\multirow{1}{*}{\rotatebox[origin=c]{0}{$Ar\rightarrow Pr$}}}
  &28.62$\pm$1.80  &30.93$\pm$1.49 &30.63$\pm$1.21 &20.33$\pm$1.50 &28.70$\pm$1.05 &29.59$\pm$0.47 &31.69$\pm$0.79 &31.83$\pm$1.08 &\textbf{32.31$\pm$1.08}\tabularnewline
% \midrule

 \parbox[t]{17mm}{\multirow{1}{*}{\rotatebox[origin=c]{0}{$Ar\rightarrow Rw$ }}}
&34.30$\pm$2.13 &35.54$\pm$1.04 &35.61$\pm$1.00 &40.85$\pm$1.56 &34.11$\pm$1.60 &34.23$\pm$0.79 &\textbf{36.77$\pm$1.54} &35.54$\pm$1.02 &35.78$\pm$1.02\tabularnewline

% \midrule

\parbox[t]{17mm}{\multirow{1}{*}{\rotatebox[origin=c]{0}{$CI\rightarrow Ar$}}}
 &18.95$\pm$1.38 &20.75$\pm$1.16 &20.79$\pm$1.76 &18.37$\pm$1.24 &23.29$\pm$1.83 &24.34$\pm$1.92  &21.34$\pm$1.57 &\textbf{23.97$\pm$1.31} &23.64$\pm$1.56\tabularnewline

% \midrule
\parbox[t]{17mm}{\multirow{1}{*}{\rotatebox[origin=c]{0}{$CI\rightarrow Pr$}}}
 &23.91$\pm$1.53 &24.70$\pm$1.47 &24.38$\pm$1.43 &22.64$\pm$1.42 &26.43$\pm$1.14 &\textbf{28.12$\pm$1.26} &25.69$\pm$1.02  &27.17$\pm$1.13 &27.46$\pm$1.46\tabularnewline

% \midrule
\parbox[t]{17mm}{\multirow{1}{*}{\rotatebox[origin=c]{0}{$CI\rightarrow Rw$}}}
&25.16$\pm$1.35 &26.08$\pm$1.29 &26.01$\pm$0.67 &22.74$\pm$1.59 &28.62$\pm$1.58 &28.86$\pm$1.02  &27.93$\pm$1.32 &29.82$\pm$1.15 &\textbf{30.17$\pm$1.23}\tabularnewline

% \midrule
% \parbox[t]{17mm}{\multirow{1}{*}{\rotatebox[origin=c]{0}{RDA}}}
% &$N$   & $N$ & $N$ & $N$ & $N$  & $N$ & $N$\tabularnewline
% \hline

% \parbox[t]{17mm}{\multirow{1}{*}{\rotatebox[origin=c]{0}{Butterfly}}}
% &$N$   & $N$ & $N$ & $N$ & $N$  & $N$ & $N$\tabularnewline

% \midrule
\parbox[t]{15mm}{\multirow{1}{*}{\rotatebox[origin=c]{0}{$Pr\rightarrow Ar$}}}
&21.79$\pm$1.92 &23.33$\pm$1.67 &24.30$\pm$1.53 &20.47$\pm$1.19 &25.37$\pm$1.28 &25.16$\pm$1.14 &24.19$\pm$1.46  &25.98$\pm$0.77 &\textbf{26.68$\pm$1.23}\tabularnewline
% \midrule
\parbox[t]{15mm}{\multirow{1}{*}{\rotatebox[origin=c]{0}{$Pr\rightarrow CI$ }}}
&16.68$\pm$1.45 &17.76$\pm$1.54 &\textbf{17.92$\pm$1.12} &15.53$\pm$1.76 &17.23$\pm$1.18 &16.53$\pm$1.94 &17.85$\pm$1.86 &16.94$\pm$1.33 &16.87$\pm$1.30\tabularnewline

\parbox[t]{15mm}{\multirow{1}{*}{\rotatebox[origin=c]{0}{$Pr\rightarrow Rw$ }}}
&33.27$\pm$1.22 &36.02$\pm$1.32 &35.34$\pm$1.39 &26.79$\pm$1.52 &35.62$\pm$1.42 &36.99$\pm$1.08  &37.43$\pm$1.24 &36.51$\pm$1.38 &\textbf{38.51$\pm$1.72}\tabularnewline

\parbox[t]{15mm}{\multirow{1}{*}{\rotatebox[origin=c]{0}{$Rw\rightarrow Ar$}}}
&29.79$\pm$1.82 &32.37$\pm$1.22 &32.11$\pm$1.58 &27.34$\pm$1.73 &32.29$\pm$1.72 &31.20$\pm$0.94  &\textbf{33.27$\pm$1.27} &32.44$\pm$1.68 &32.36$\pm$1.17\tabularnewline

\parbox[t]{15mm}{\multirow{1}{*}{\rotatebox[origin=c]{0}{$Rw\rightarrow CI$}}}
&19.71$\pm$1.85 &20.90$\pm$1.54 &20.43$\pm$0.92 &18.47$\pm$1.61 &20.58$\pm$1.17 &19.93$\pm$1.91  &20.79$\pm$0.94 &19.61$\pm$1.14 &\textbf{21.82$\pm$1.19}\tabularnewline

\parbox[t]{15mm}{\multirow{1}{*}{\rotatebox[origin=c]{0}{$Rw\rightarrow Pr$}}}
&37.29$\pm$1.99 &26.28$\pm$1.44 &39.42$\pm$1.55 &31.61$\pm$1.65 &37.88$\pm$1.04 &40.10$\pm$1.46  &28.93$\pm$1.38 &38.99$\pm$1.66 &\textbf{42.50$\pm$1.39}\tabularnewline
\midrule
\parbox[t]{15mm}{\multirow{1}{*}{\rotatebox[origin=c]{0}{Average}}}
&25.46$\pm$1.71 &26.02$\pm$1.41 &27.02$\pm$1.29 &23.33$\pm$1.56 &27.27$\pm$1.30 &27.68$\pm$1.36 &27.05$\pm$1.35 &28.03$\pm$1.15 &\textbf{28.83$\pm$1.24}
\tabularnewline
\bottomrule
\end{tabular}}}
\caption{Target accuracy (\%) on Office-Home datasets with Unif-40\% noise. The best results are highlighted in bold.}
\label{home2}
\par\end{center}
\end{table*}

\section{C. Definition of Noise}
We manually inject two types of noisy labels into the source domain by the label transition matrix Q, where $Q_{ij}=Pr(\tilde{y}^s=j|y^s=i)$: (1) \textbf{Uniform noise}. The ground-truth label of each sample can change to any wrong class with probability $\frac{\rho}{K}$ independently, while the label still have probability $1-\frac{1-K}{K}\rho$ to be correct. (2) \textbf{Flip noise}. We consider the case that the ground-truth label of each sample can flip to one similar class with probability of $\rho$, where $\rho$ denotes the noise rate, and K denotes the the dimension of the label space. 
 
 Concretely, $Q_U$ and $Q_F$ representing the label transition matrix of Uniform noise and Flip noise respectively are shown as follows.

$$Q_U=
\left[
\begin{matrix}
 1-\frac{1-K}{K}\rho     & \frac{\rho}{K}      & \cdots &\frac{\rho}{K}     \\
 \frac{\rho}{K}      & 1-\frac{1-K}{K}\rho      & \cdots & \frac{\rho}{K}      \\
 \vdots & \vdots & \ddots & \vdots \\
 \frac{\rho}{K}      & \frac{\rho}{K}      & \cdots & 1-\frac{1-K}{K}\rho        \\
\end{matrix}
\right]_{K\times K}
$$

$$Q_F=
\left[
\begin{matrix}
 1-\rho     & 0      & \cdots &\rho     \\
 \rho       & 1-\rho   & \cdots     & 0      \\
 \vdots & \vdots & \ddots & \vdots \\
 \rho      & 0      & \cdots & 1-\rho        \\
\end{matrix}
\right]_{K\times K}
$$

\section{D. Additional experiments on FixBi}

Fixi \cite{na2021fixbi} is an unsupervised domain adaptation method, which generates multiple intermediate domains via a fixed radio-based mixup to decrease the discrepancy between the source domain and the target domain. 
\begin{table}[htb]
    \centering
    \setlength{\tabcolsep}{6pt}{
    \renewcommand{\arraystretch}{1}
    \resizebox{0.46\textwidth}{!}{
    \begin{tabular}{c|cccc}
    \toprule 
         &$D\rightarrow W$ &$W\rightarrow D$ &$A\rightarrow D$ &$A\rightarrow W$\tabularnewline
         \midrule
         FixBi &44.13 &48.59 &32.66 &35.46\tabularnewline
         \midrule
         $\text{GearNet}_\text{FixBi}$ &\textbf{63.39} &\textbf{70.56} &\textbf{73.99} &\textbf{42.86} \tabularnewline
         \midrule
    \end{tabular}}}
    \caption{Target accuracy (\%) on various tasks with Unif-20\% noise. Bold numbers are superior results.}
    \label{review3}
\end{table}

To verify that our method can improve FixBi, We conduct the additional experiments on four tasks from Office-31 with Unif-20\% noise and the results are shown in Table \ref{review3}. Since FixBi is designed for unsupervised domain adaptation, it performs poorly when the source domains are with noisy labels. After equipped with our method, the performance is significantly improved by more than 40\%, thereby explicitly illustrating the effectiveness of our method.

\bibliography{neurips_2021}